\documentclass{article}

\usepackage{fullpage,latexsym}
\usepackage{epsfig}
\usepackage{pslatex}

\usepackage{amsmath,epsf}
\numberwithin{equation}{section} % in amsmath

 \newtheorem{lemma}{Lemma}[section]
 \newtheorem{theorem}[lemma]{Theorem}

 \newtheorem{definition}[lemma]{Definition}
 \newtheorem{rem}[lemma]{Remark}
\newenvironment{remark}{\begin{rem}}{\hspace*{\fill}$\diamondsuit$\end{rem}}
\newenvironment{proof}{\par \sc Proof.\rm}{\hspace*{\fill}$\Box$\vspace{1ex}}
 \newtheorem{ex}[lemma]{Example}
\newenvironment{example}{\begin{ex}}{\hspace*{\fill}$\diamondsuit$\end{ex}}

\newcommand{\commentout}[1]{}

\begin{document}

\title{Clustering by Compression}
\author{Rudi Cilibrasi\thanks{Supported in part by NWO.
Address:
CWI, Kruislaan 413,
1098 SJ Amsterdam, The Netherlands.
Email: {\tt Rudi.Cilibrasi@cwi.nl}.}
\\CWI
\and
Paul Vitanyi\thanks{Supported in part by the
EU project RESQ, IST--2001--37559, the NoE QUIPROCONE
+IST--1999--29064,
the ESF QiT Programmme, and the EU Fourth Framework BRA
 NeuroCOLT II Working Group
EP 27150. 
Address:
CWI, Kruislaan 413,
1098 SJ Amsterdam, The Netherlands.
Email: {\tt Paul.Vitanyi@cwi.nl}.}\\CWI and University of Amsterdam
%\and
%Ronald de Wolf\thanks{Supported in part by EU project RESQ, IST-2001-37559.
%Address:
%CWI, Kruislaan 413,
%1098 SJ Amsterdam, The Netherlands.
%Email: {\tt Ronald.de.Wolf@cwi.nl}.}\\
%CWI
}
\date{}
\maketitle

\begin{abstract}
We present a new method for clustering based on compression.
The method 
doesn't use subject-specific features or background knowledge, 
and works as follows:
First, we determine a universal similarity 
distance, the normalized compression
distance or NCD,
computed from the lengths of compressed data files 
(singly and in pairwise concatenation).
Second, we apply a hierarchical clustering method.
The NCD is universal in that it is
not restricted to a specific application area, and
works across application area boundaries.
A theoretical precursor, the normalized information distance,
co-developed by one of the authors,
is provably optimal in the sense that it minorizes 
every computable normalized metric that satisfies a certain density requirement.
However, the optimality comes at the price
of using the non-computable notion of Kolmogorov complexity.
We propose precise notions of similarity metric, normal compressor, and
show that the NCD based on a normal compressor is a similarity metric 
that approximates optimality.
To extract a hierarchy of clusters
from the distance matrix,
we determine a dendrogram (binary tree) 
by a new quartet
method and a fast heuristic to implement it.
The method is implemented and available as public software, and is
robust under choice of different compressors.
To substantiate our claims of universality and robustness,
we report evidence of successful application in areas as diverse as
genomics, virology, languages, literature, music, handwritten digits,
astronomy, and 
combinations of objects from completely different
domains, using statistical, dictionary, and block sorting compressors.
In genomics we presented new evidence for major questions
in Mammalian evolution, based on whole-mitochondrial genomic
analysis: the Eutherian orders and the Marsupionta hypothesis
against the Theria hypothesis.
\end{abstract}

\section{Introduction}
\label{sect.intro}

All data are created equal but some data are more alike than others.
We propose a method expressing this alikeness,
using a new similarity metric based on compression.
This metric 
doesn't use any features or background knowledge, and can without
changes be applied to different areas and across area boundaries.
It is robust in the sense that its success appears independent
from the type of compressor used.
The clustering we use is hierarchical clustering in dendrograms
based on a new fast heuristic for the quartet method. 
The method is available as an open-source software tool. 
%In a sense, the feature-freeness of our 
%compression scheme is a property shared
%with genome alignment to extract a similarity distance matrix
%in computational genomics. But in the latter case one commonly selects
%parts of the genomes that are suspected of being important to
%do the alignment on. This is a form of using background knowledge.
%Our proposed method gives satisfactory results in all of the many
%areas we have tested it. The results are commonly less precise than
%those obtained using the common special-purpose methods optimized
%and supplied with background knowledge about the targeted area or problem.
%But the redeeming feature of our method is that it is 
%perfectly universal, and can be applied to
%all areas and across area boundaries without change. 
Below we explain the method, the theory underpinning it, 
and present evidence for its universality and robustness by 
experiments and results in a plethora of different areas using
different types of compressors.

{\bf Feature-Based Similarities:}
We are presented with unknown data and
the question is to determine the similarities among them
and group like with like together. Commonly, the data are
of a certain type: music files, transaction records of ATM machines,
credit card applications, genomic data. In these data there are
hidden relations that we would like to get out in the open.
For example, from genomic data one can extract
letter- or block frequencies (the blocks are over the four-letter alphabet);
 from music files one can extract 
various specific numerical features,
related to pitch, rhythm, harmony etc.
One can extract such features using for instance
Fourier transforms~\cite{TC02} or wavelet transforms~\cite{GKCwavelet}.   
The feature vectors corresponding to the various files are then
classified or clustered using existing classification software, based on
various standard statistical pattern recognition classifiers~\cite{TC02},
Bayesian classifiers~\cite{DTWml},
hidden Markov models~\cite{CVfolk},
ensembles of nearest-neighbor classifiers~\cite{GKCwavelet}
or neural networks~\cite{DTWml,Sneural}.
For example, in music one feature would be to look for rhythm in the sense
of beats per minute. One can make a histogram where each histogram
bin corresponds to a particular tempo in beats-per-minute and
the associated peak shows how frequent and strong that
particular periodicity was over the entire piece. In \cite{TC02}
we see a gradual change from a few high peaks to many low and spread-out
ones going from hip-hip, rock, jazz, to classical. One can use this
similarity type to try to cluster pieces in these categories.
However, such a method requires specific and detailed knowledge of
the problem area, since one needs to know what features to look for.
                                                                                
{\bf Non-Feature Similarities:} Our aim 
is to capture, in a single similarity metric,
{\em every effective metric\/}:
effective versions of Hamming distance, Euclidean distance,
edit distances, alignment distance, Lempel-Ziv distance \cite{CPSV00}, 
and so on.
This metric should be so general that it works in every
domain: music, text, literature, programs, genomes, executables,
natural language determination,
equally and simultaneously.
It would be able to simultaneously detect {\em all\/}
similarities between pieces that other effective metrics can detect.

{\bf Compression-based Similarity:}
Such a ``universal'' metric 
was co-developed by us in \cite{LBCKKZ01,Li01,Li03}, as a normalized
version of the ``information metric'' of \cite{LiVi97,BGLVZ98}.
Roughly speaking, two objects are deemed close if
we can significantly ``compress'' one given the information
in the other, the idea being that if two pieces are more similar,
then we can more succinctly describe one given the other.
The mathematics used is based on Kolmogorov complexity theory \cite{LiVi97}.
In \cite{Li03} we defined a
new class of metrics, taking values in $[0,1]$ and
appropriate for measuring effective
similarity relations between sequences, say one type of similarity
per metric, and {\em vice versa}. It was shown that an appropriately
``normalized'' information distance
minorizes every metric
in the class. 
It discovers all effective similarities in the sense that if two
objects are close according to some effective similarity, then
they are also close according to the normalized information distance.
Put differently, the normalized information distance represents
similarity according to the dominating shared feature between
the two objects being compared.
The normalized information distance too is a metric 
and takes values in $[0,1]$;
hence it may be called {\em ``the'' similarity metric}.
To apply this ideal precise mathematical theory in real life, 
we have to replace the use of  the noncomputable
Kolmogorov complexity by an approximation 
using a standard real-world compressor. 
Earlier approaches resulted in
the first completely automatic construction
of the phylogeny tree based on whole mitochondrial genomes,
\cite{LBCKKZ01,Li01,Li03}, 
a completely automatic construction of a language tree for over 50
Euro-Asian languages \cite{Li03}, 
detects plagiarism in student programming assignments 
\cite{SID}, gives phylogeny of chain letters \cite{BLM03}, and clusters
music \cite{CVW03}.
Moreover, the method turns out to be robust under change of the underlying
compressor-types: statistical (PPMZ), Lempel-Ziv based  dictionary (gzip),
block based (bzip2), or special purpose (Gencompress).

{\bf Related Work:}
In view of the simplicity and naturalness of our proposal, 
it is perhaps surprising that
compression based clustering and classification
approaches did not arise before. But recently there have been
several partially independent proposals in that direction:
\cite{BCL02a,BCL02b} for building language trees---while 
citing \cite{LiVi97,BGLVZ98}---is by essentially more {\em ad hoc\/} arguments 
about empirical Shannon entropy and Kullback-Leibler distance. 
This approach is used
to cluster music MIDI files by Kohonen maps in \cite{LLB03}.   
Another recent offshoot based on our work is \cite{KSAG03}
hierarchical clustering based on mutual information.
In a related, but considerably simpler feature-based approach, one can compare 
the word frequencies in text files to assess similarity.
In \cite{YPYG03} the word frequencies of words common to a pair
of text files are used as entries in two vectors, and the similarity
of the two files is based on the distance between those vectors.
The authors attribute authorship to Shakespeare plays, the Federalist
Papers, and the Chinese classic ``The Dream of the Red Chamber.''
The approach to similarity distances based on block occurrence
statistics is standard in genomics, and in an experiment below
it gives inferior phylogeny trees compared  to our compression method
(and wrong ones according to current biological wisdom).
The possibly new feature in the cited work is that it uses
statistics of only the words that the files being compared have in common. 
A related, opposite, approach was taken in \cite{KAS03},
where literary texts are clustered by author gender or fact versus
fiction, essentially by first identifying distinguishing features,
like gender dependent word usage,
and then classifying according to those features.

{\bf Outline:}
Here we propose a first comprehensive theory 
of real-world compressor-based normalized compression distance, 
a novel hierarchical clustering heuristic, together with many new 
applications. First,
we define new mathematical notions of ``similarity metric,'' ``normal
compressor,'' and ``normalized compression
distance.'' We then prove the normalized compression
distance based on a normal compressor to be a similarity metric.
The normalized compression
distance is shown to be quasi-universal in the sense that it minorizes
every computable similarity metric up to an additive error term 
that depends on the quality of the compressor's approximation of the true
Kolmogorov complexities of the files concerned, and that under certain
conditions vanishes with increasing file length.    
This means that the NCD captures the dominant similarity over all 
possible features of the objects compared, up to the stated precision.
Next, we present a method of hierarchical clustering based on 
a novel fast randomized hill-climbing heuristic of a new quartet tree
optimization criterion. Given a matrix of the pairwise similarity
distances between the objects, we score how well the resulting tree
represents the information in the distance matrix on a scale of 0 to 1.
Then, as proof of principle, we run the program on three
data sets, where we know what 
the final answer should be: (i) reconstruct a tree from a distance
matrix obtained from a randomly generated tree; (ii)
reconstruct a tree from files containing artificial similarities;
and (iii) reconstruct a tree from natural files of vastly different types.
To substantiate our claim of universality,
we apply the method to different areas,
not using any feature analysis
at all.
We first give an example in whole-genome phylogeny
using the whole mitochondrial DNA of the species concerned.
We compare the hierarchical clustering of our method with
a more standard method of two-dimensional clustering (to show
that our dendrogram method of depicting the clusters is more informative). 
We give a whole-genome phylogeny of fungi and compare
this to results using alignment of selected proteins (alignment being often
too costly to perform on the whole-mitochondial genome, but the disadvantage
of protein selection being that different selections usually
result in different phylogenies---so which is right?). We identify
the virii that are closest to the sequenced SARS virus;
we give an example of clustering of language families; Russian authors
in the original Russian, the same pieces in English translation
(clustering partially follows the translators); clustering of music
in MIDI format; clustering of handwritten digits used for optical
character recognition; and clustering of radio observations of a 
mysterious
astronomical object, a microquasar of extremely complex variability. 
In all these cases the method performs very well in the following
sense: The method yields the phylogeny of 24 species precisely
according to biological wisdom. The probability that it randomly would
hit this one outcome, or anything reasonably close,
is very small. In clustering 36 music
pieces taken equally many from pop, jazz, classic, so that 12-12-12 
is the grouping
we understand is correct,
we can identify convex clusters so that only six errors are made. 
(That is, if three items get dislodged then six items get 
misplaced.)
The probability that this happens by chance is extremely small.
The reason why we think the method does something remarkable
is concisely put by Laplace \cite{La51}:
\begin{quote}
``If we seek a cause wherever we perceive symmetry,
it is not that we regard the symmetrical event as less
possible than the others, but, since this event ought to
be the effect of a regular cause or that of chance,
the first of these suppositions is more probable
than the second. On a table we see letters arranged in
this order
{\tt C o n s t a n t i n o p l e},
and we judge
that this arrangement is not the result of chance,
not because it is less possible than others, for
if this word were not employed in any language we
would not suspect it came from any particular cause, but
this word being in use among us, it is incomparably
more probable that some person has thus arranged the aforesaid letters
than that this arrangement is due to chance.''
\end{quote}

{\bf Materials and Methods:}
The data samples we used were obtained from standard data bases
accessible on the world-wide web, generated by ourselves,
or obtained from research groups in the field of investigation.
We supply the details with each experiment. The method of processing
the data was the same in all experiments.
First, we preprocessed the data samples to bring 
them in appropriate format: the genomic material over the four-letter
alphabet $\{A,T,G,C\}$ is recoded in a four-letter alphabet;
the music MIDI files are stripped of identifying information
such as composer and name of the music piece.
Then, in all cases the data samples were completely automatically
processed by our CompLearn Toolkit, rather than as is usual in
phylogeny, by using an ecclectic set of software tools per 
experiment.
Oblivious to the problem area concerned, simply using the distances
according to the NCD below, 
the method described in this paper fully automatically
classifies the objects concerned.
The method has been released in the public domain as open-source software
at http://complearn.sourceforge.net/   .
The CompLearn Toolkit is a suite
of simple utilities that one can use to apply compression
techniques to the process of discovering and learning patterns
%The compression-based approach used is powerful because it 
%can mine patterns in
in completely different domains. 
%It can cluster musical styles of pieces of music
%and unknown composers. 
%It can cluster the languages of bodies of text, or cluster the
%text by authors in the same language.
%It can discover the relationships between species
%and the origin of new unknown viruses such as SARS. 
In fact, this method is so general that it requires 
no background knowledge about any particular 
subject area. There are no domain-specific parameters to set,
and only a handful of general settings.

The Complearn Toolkit using NCD and not, say, alignment,
can cope with full genomes and 
other large data files and thus comes up with a single
distance matrix. The clustering heuristic generates a tree
with a certain confidence, called standardized benefit score or $S(T)$ value
in the sequel. Generating trees from the same distance
matrix many times resulted in the same tree or almost the same tree,
for all distance matrices we used, even though the heuristic is randomized.
The differences that arose are apparently due to early or late termination
with different $S(T)$ values.
This is a great difference with previous phylogeny methods, where because of
computational limitations one uses only parts of the genome, or certain 
proteins that are viewed as significant \cite{KBSMJ01}. These are
run through a tree reconstruction method like neighbor joining \cite{SN87},
maximum likelihood, maximum evolution, maximum parsimony as in \cite{KBSMJ01},
or quartet hypercleaning \cite{Br00},
many times. The percentage-wise agreement on certain branches arising
are called ``bootstrap values.'' Trees are depicted with the best bootstrap
values on the branches that are viewed as supporting the theory tested.
Different choices of proteins result in different best trees. One way
to avoid this ambiguity is to use the full genome,
\cite{RWKC03,Li03}, leading to whole-genome phylogeny. With our
method we do whole-genome phylogeny, and end up with a single
overall best tree, not optimizing selected parts of it.

The quality of the results depends on (a) the NCD distance matrix,
and (b) how well the hierarchical tree represents the information
in the matrix. The quality of (b) is measured by the $S(T)$ value,
and is given with each experiment.
In general, the $S(T)$ value deteriorates
for large sets. We believe this to be partially an
artifact of a low-resolution  NCD 
matrix due to limited compression power, and limited file size.
The main reason, however, is the fact that with increasing 
size of 
a natural
data set 
the projection of the information 
in the NCD matrix into a
binary tree gets increasingly distorted.
Another aspect limiting the quality of the NCD matrix is more subtle. 
Recall that the method knows nothing
about any of the areas
we apply it to. It determines the dominant feature as seen through
the NCD filter. The dominant feature of alikeness between
two files may not correspond to our a priori conception but may have
an unexpected cause. The results of our experiments suggest that this
is not often the case: In the natural data sets where we have
preconceptions of the outcome, for example that works by the same 
authors should cluster together, or music pieces by the same composers,
musical genres, or genomes, the outcomes conform largely to our expectations.
For example, in the music genre experiment the method would fail
dramatically if genres were evenly mixed, or mixed with little bias.
However, to the contrary, the separation in clusters is almost perfect.
The few misplacements that are discernable are either errors (the method
was not powerful enough to discern the dominant feature), or the dominant
feature between a pair of music pieces is not the genre but some
other aspect. The surprising news is that we can generally confirm
expectations with few misplacements, indeed, that the data don't
contain unknown rogue features that dominate to cause spurious
(in our preconceived idea) clustering. This gives evidence that where the
preconception is in doubt, like with phylogeny hypotheses, the 
clustering can give true support of one hypothesis against another one.

{\bf Figures:}
We use two styles to display the hierarchical clusters.
In the case of genomics of Eutherian orders and fungi,
language trees, 
it is convenient to follow the
dendrograms that are customary in that area (suggesting temporal
evolution) for easy comparison with the literature. 
Although there is no temporal relation intended, the dendrogram
representation looked also appropriate for the
Russian writers, and translations of Russian writers.
In the other experiments (even the genomic SARS experiment) it is more informative to 
display an unrooted ternary tree (or binary tree if we think about
incoming and outgoing edges) with explicit internal nodes. 
This facilitates identification of clusters in terms of
subtrees rooted at internal nodes or contiguous sets of subtrees rooted
at branches of internal nodes.

\section{Similarity Metric}

In mathematics, different distances arise in all sorts of contexts,
and one usually requires these to be a ``metric''.
We give a precise formal meaning to the loose
distance notion of ``degree of similarity''
used in the pattern recognition
literature.

{\bf Metric:} 
Let $\Omega$ be a nonempty set and ${\cal R}^+$ be the set of nonnegative
real numbers.
A {\em metric} on $\Omega$ is a function 
$D: \Omega \times \Omega \rightarrow {\cal R}^+$ 
satisfying the metric (in)equalities:
\begin{itemize}
\item
$D(x,y)=0$ iff $x=y$, 
\item
$D(x,y)=D(y,x)$ (symmetry), and
\item
$D(x,y)\leq D(x,z)+D(z,y)$ (triangle inequality).
\end{itemize}
The value $D(x,y)$ is called the {\em distance} between $x,y \in \Omega$.
A familiar example of a metric is the Euclidean metric, 
the everyday distance $e(a,b)$ between two objects $a,b$ 
expressed in, say, meters.  
Clearly, this distance satisfies the properties
$e(a,a)=0$, $e(a,b)=e(b,a)$, and $e(a,b) \leq e(a,c) + e(c,b)$
(for instance, $a=$ Amsterdam, $b=$ Brussels, and $c=$ Chicago.) 
We are interested in ``similarity metrics''. 
For example, if the objects are classical music pieces
then the function $D(a,b)= 0$ if $a$ and $b$ are by the same composer
and $D(a,b) = 1$ otherwise,
is a similarity metric.
This metric captures only one similarity aspect (feature) 
of music pieces, presumably an important one because it subsumes
a conglomerate of more elementary features.

{\bf Density:}
In defining a class of acceptable metrics
we want to exclude unrealistic metrics
like $f(x,y) = \frac{1}{2}$ for {\em every} pair $x  \neq y$.
We do this by restricting
the number of objects within a given distance of an object.
As in \cite{BGLVZ98} we do this by only considering effective distances,
as follows.
Fix a suitable, and for the remainder of the paper, fixed, programming
language. This is the {\em reference programming} language. 

\begin{definition}\label{def.em}
\rm
Let $\Omega = \Sigma^*$, with $\Sigma$ a finite nonempty alphabet
and $\Sigma^*$ the set of finite strings over that alphabet. Note that
for us ``files'' in computer memory are finite binary strings.
A function $D: \Omega \times \Omega \rightarrow {\cal R}^+$ is an 
{\em acceptable metric} if for every pair of objects $x,y \in \Omega$
the distance  $D(x,y)$ is 
the length of a binary prefix code-word that is a program that computes 
$x$ from $y$, and vice versa, in the reference programming language,
and the metric (in)equalities hold up to $O(\log n)$ where $n$
is the maximal binary length of an element of $\Omega$ involved in
the (in)equality concerned.
\end{definition}

\begin{example}\label{ex.ham}
\rm
In representing the Hamming distance $d$ between 
$x$ and $y$ strings of equal length $n$ differing in positions 
$i_1, \ldots , i_d$, we can use a simple 
prefix-free encoding of $(n,d,i_1, \ldots , i_d)$ 
in $H(x,y)=2 \log n + 4 \log \log n +2 + d \log n$ bits.
We encode $n$ and $d$ prefix-free in $\log n+2 \log \log n +1$
bits each, see e.g. \cite{LiVi97},
and then the literal indexes of the actual flipped-bit 
positions. Hence, $H(x,y)$ is the length of a prefix code word (prefix
program) to compute $x$ from $y$ and {\em vice versa}.
Then, by the Kraft inequality, see \cite{CT91},
\begin{equation}\label{eq.dc}
\sum_{y \neq x, |y|=|x|} 2^{-H(x,y)} \leq 1.
\end{equation}
\end{example}
It is easy to verify that $H$ is a metric in the sense that it satisfies
the metric (in)equalities up to $O(\log n)$
additive precision.

{\bf Normalization:}
Large objects (in the sense of long strings)
that differ by a tiny part are intuitively
closer than tiny objects that differ by the same amount.
For example, two whole mitochondrial genomes 
of 18,000 bases that differ by 9,000 are very different, while two whole
nuclear genomes of $3 \times 10^9$ bases
 that differ by only 9,000 bases are very similar.
Thus, absolute difference between two objects doesn't govern similarity,
but relative difference appears to do so.
\begin{definition}
\rm
A {\em compressor} is a lossless encoder mapping $\Sigma^*$ into
$\{0,1\}^*$ such that the resulting code is a prefix code. 
For convenience of notation we identify ``compressor''
with a ``code word length function'' $C: \Sigma^* \rightarrow {\cal N}$,
where ${\cal N}$ is the set of nonnegative integers.
The compressed
version of a file $x$ is denoted by $x^*$ and its length is $C(x) = |x^*|$.
We only consider compressors such that $C(x) \leq |x|+O(\log |x|$.
\end{definition}
Since the compressor is a lossless encoder, 
if $x \neq y$ then $x^* \neq y^*$.
In the following we fix a compressor $C$, at this stage it
doesn't matter which one.
We call the fixed compressor
the {\em reference compressor}.
%Relative to this compressor, we normalize a metric as follows:
%Let $D(x,y)$ be a
%distance satisfying the density condition (\ref{eq.dc}).
%Let $norm$ be a function such that  $d(x,y) = norm(D,x,y)$ has values in
\begin{definition}\label{eq.defsm}
\rm
A {\em normalized metric} or {\em similarity metric},
relative to a reference compressor $C$,
is a function $d: \Omega \times \Omega \rightarrow [0,1]$ 
that for every constant $e \in [0,1]$ satisfies the density constraint 
\begin{equation}\label{eq.dp}
 | \{(x,y): d(x,y) \leq e \leq 1, C(y) \leq C(x) \} | < 2^{(1+e) C(x)+1} ,
\end{equation}
and satisfies the metric (in)equalities up to additive precision
$O((\log n)/n)$ where $n$ is the maximal binary length of an element 
of $\Omega$ involved in the (in)equality concerned.
\end{definition}
This requirement follows from a ``normalized'' version 
of the Kraft inequality:
\begin{lemma}\label{lem.ki}
Let $d: \Omega \times \Omega \rightarrow [0,1]$ satisfy
\begin{equation}\label{eq.nc}
\sum_{y \neq x} 2^{- (1+d(x,y))C(x)} \leq 1 .
\end{equation}
Then,  $d$  satisfies \eqref{eq.dp}.
\end{lemma}
\begin{proof}
For suppose the contrary: 
there is an $e \in [0,1]$, such that
\eqref{eq.dp} is false. Then,
starting from (\ref{eq.nc}) we obtain
a contradiction:
$$
1 \geq \sum_{y \neq x} 2^{-(1+d(x,y)) C(x)} 
\geq  \sum_{y \neq x \& d(x,y) \leq e \leq 1  \& C(y) \leq C(x)}
2^{-(1+e)C(x)} 
 \geq 2^{(1+e) C(x)+1} 2^{-(1+e)C(x)} > 1.
$$

\end{proof}

We call a normalized metric a ``similarity'' metric, because
it gives a relative similarity according to the metric distance
(with distance 0 when objects are maximally similar and distance 1 when
the are maximally dissimilar)
and, conversely, for every well-defined computable notion of similarity we
can express it as a metric distance according to our definition.
In the literature a distance that expresses lack of similarity (like
ours)  is often
called a ``dissimilarity'' distance or a ``disparity'' distance.

\begin{remark}
\rm
As far as the
authors know, the idea of normalized metric is, surprisingly, not
well-studied. An exception is \cite{Ya02}, which investigates
normalized metrics 
to account for relative
distances rather than absolute ones, and it does so
 for much the same reasons as in
the present work. An example there is the normalized  
Euclidean metric $|x-y|/(|x|+|y|)$, where $x,y \in {\cal R}^n$
(${\cal R}$ denotes the real numbers)  and $| \cdot |$ is the
Euclidean metric---the $L_2$ norm. Another example is a
normalized symmetric-set-difference metric.
But these normalized metrics are not necessarily effective in that
the distance between two objects gives the length of an effective
description to go from either object to the other one.
\end{remark}

\begin{example}
\rm
The prefix-code for the Hamming distance $H(x,y)$ between 
$x,y \in \{0,1\}^n$ in
Example~\ref{ex.ham} is a program to compute from $x$ to $y$ and
{\em vice versa}. It is metric up to $O(\log n)$ additive
precision. To turn it into a similarity metric define $h(x,y)=H(x,y)/C_n$,
where $C_n = \max \{C(x): |x|=n\}$.
Trivially, $0 \leq h(x,y) \leq 1$.
The metric properties of $H(x,y)$ that held up to
additive precision $O(\log n)$ are preserved under division by $C_n$ with
the precision improved to $O((\log n) /n)$. Since $C(x)/C_n \leq 1$
we have $\sum_{y:y \neq x, |y|=|x|=n} 2^{-h(x,y)C(x)} \leq 1$.
Then, $\sum_{y \neq x, |y|=|x|=n} 2^{-(1+h(x,y))C(x)} 
\leq \sum_{|x|=n} 2^{-C(x)} \leq 1$, where the last inequality
is the Kraft inequality since $C(x)$ is a prefix code word
length.  Hence $h(x,y)$ satisfies \eqref{eq.nc} and therefore
\eqref{eq.dp} by Lemma~\ref{lem.ki}.
\end{example}

\section{Normalized Compression Distance}
                                                                                
In \cite{BGLVZ98,Li03} the conditions and definitions on a similarity
metric are the particular instance where the compressor $C$ is 
instantiated as the ultimate
compressor $K$ such that $K(x)$ is the Kolmogorov complexity
\cite{LiVi97} of $x$.
In \cite{Li03}, it is shown that one can
represent the entire wide class of the similarity metrics 
(the class based on the ultimate compressor $K$)
by a single representative:
the ``normalized information distance'' is a metric, and it is 
universal in the sense that this single metric uncovers all similarities 
simultaneously that the various metrics in the class uncover separately.
This should be understood in the sense that if two files (of whatever type)
are similar (that is, close) according to the particular feature described by 
a particular metric, then they are also similar (that is, close)
in the sense of the normalized information metric. This justifies
calling the latter {\em the\/} similarity metric.
However, this metric is based on the 
notion of Kolmogorov complexity. The Kolmogorov complexity of a file
is essentially the length of the ultimate compressed version of the file. 
Unfortunately, the Kolmogorov complexity 
of a file is non-computable in the 
Turing sense. 
In applying
the approach, we have to make do with an approximation based on a 
far less powerful real-world reference compressor $C$.
The resulting applied approximation of the ``normalized
information distance'' is called the 
{\em normalized compression distance (NCD)}
and is defined by
\begin{equation}\label{eq.ncd}
NCD(x,y) = \frac{C(xy)- \min \{C(x),C(y)\}}{ \max\{C(x),C(y)\}}.
\end{equation}
Here,
$C(xy)$ denotes the compressed size of the concatenation of $x$ and $y$,
$C(x)$ denotes the compressed size of $x$,
and $C(y)$ denotes the compressed size of $y$.
The NCD is a non-negative number $0 \leq  r \leq 1 + \epsilon$ representing how 
different the two files are. Smaller numbers represent more similar files. 
The $\epsilon$ in the upper bound is due to 
imperfections in our compression techniques, 
but for most standard compression algorithms one is unlikely 
to see an $\epsilon$ above 0.1 (in our experiments gzip and bzip2 achieved
NCD's above 1, but PPMZ always had NCD at most 1).

\begin{remark}\label{rem.compensate}
\rm
Technically, the {\em  Kolmogorov complexity} of $x$ given $y$ is the length
of the shortest binary program that on input $y$ outputs $x$;
it is denoted as $K(x|y)$. For precise definitions, theory and applications,
see \cite{LiVi97}. The Kolmogorov complexity of $x$ is the length
of the shortest binary program with no input that outputs $x$;
it is denoted as $K(x)=K(x|\epsilon)$ where $\epsilon$ denotes the empty input. 
The similarity metric in \cite{Li03} is 
$\max\{K(x|y),K(y|x)\}/\max\{K(x),K(y)\}$. Approximation
of the denominator by a given compressor is straightforward
by $\max\{C(x),C(y)\}$. The numerator is more tricky. It can
be written as 
\begin{equation}\label{eq.nom}
\max \{ K(x,y)-K(x),  K(x,y)-K(y) \},
\end{equation}
 within logarithmic
additive precision by the additive property
of Kolmogorov complexity \cite{LiVi97}. The term $K(x,y)$ represents the length of the shortest
program for the pair $(x,y)$. In compression practice it is easier
to deal with the concatenation $xy$ or $yx$. Again, within logarithmic
precision $K(x,y)=K(xy)=K(yx)$. But we have to deal with a real-life
compressor here, for which $C(xy)$ may be different from $C(yx)$.
Clearly, however, the smaller of the two will be the closest approximation
to $K(x,y)$. Therefore, following a suggestion by
Steven de Rooij, one can approximate \eqref{eq.nom} best by
$\min\{C(xy),C(yx)\} - \min \{C(x),C(y)\}$. 
Here, and in the CompLearn Toolkit, however, we simply use $C(xy)$ rather than
$\min\{C(xy),C(yx)\}$. This is justified by the observation
that block-coding based compressors are symmetric 
almost by definition, and experiments with various stream-based 
compressors (gzip, PPMZ) show only small 
deviations from symmetry. In our definition
of a ``normal'' compressor below we put symmetry as one of 
the basic properties.
\end{remark}
The theory as developed for the Kolmogorov-complexity based ``normalized
information distance'' in \cite{Li03}
 does not hold directly for the (possibly poorly) approximating NCD.
Below, we develop the theory of NCD based on the 
notion of a ``normal compressor,'' and show that the NCD is a (quasi-)
universal similarity metric relative to a normal reference compressor $C$.
The theory developed in \cite{Li03} is
the boundary case $C=K$, where the ``quasi-universality''
below has become full ``universality''.

\subsection{Normal Compressor}
The properties of the NCD defined below
depend on the details of the compressor used.
However, under mild and natural assumptions on the compressor
one can show the NCD is a (quasi-)universal similarity metric.
%In practice, in some cases one or more of 
%the conditions of ``normality'' will only
%be satisfied up to a certain precision by a given compressor,
%which then is reflected in the precision of the metric properties and
%upper and lower bound on the range of the NCD. 

\begin{definition}
\rm
%Let $C$ be a compressor and let $x \in \Omega$ be a string.
%We denote the compressed binary version of $x$ by $x^*$ and its
%length by $C(x)=|x^*|$. 
A compressor $C$ is {\em normal}
if it satisfies, up to an additive $O(\log n)$ term,
with $n$ the maximal binary length of an element of $\Omega$
involved in the (in)equality concerned,
the following:
\begin{enumerate}
%\item
%{\em Stream-basedness}: 
%In the process of compressing $xy$,
%the compressor first outputs the compressed version $x^*$ of $x$
%and then outputs a compressed version of $y$, denoted by $(y|x)^*$.
%Denoting $C(y|x)=|(y|x)^*|$ we have $C(xy)=C(x)+C(y|x)$.
\item
{\em Idempotency}: $C(xx)=C(x)$.
\item
{\em Monotonicity}: $C(xy) \geq C(x)$.
\item 
{\em Symmetry}: $C(xy)=C(yx)$.
\item
{\em Distributivity}: 
$C(xy) + C(z) \leq C(xz)+C(yz)$.
\end{enumerate}
\end{definition}

%{\bf Stream-Basedness:}
%To develop a theory of real compressors, paralleling the abstract Kolmogorov
%complexity theory of the ideal similarity metric in \cite{Li03},  
%we need to assume that
%the reference compressor is a stream-based compressor
%such as the PPM family or the Lempel-Ziv family. 
%In the applications we discuss in the remainder of the paper,
%however, we used both stream-based compressors like gzip and PPMZ,
%and non-stream-based
%compressors like bzip2. The results turned out to be robust
%among change of compressors.

{\bf Idempotency:}
A reasonable compressor will see exact repetitions and obey
idempotency up to the required precision.

{\bf Monotonicity:}
A real compressor 
must have the monotonicity property, at least up to the required
precision. The property is evident for stream-based compressors, 
and only slightly less evident for block-coding compressors.

{\bf Symmetry:}
Stream-based compressors of the Lempel-Ziv family, like gzip and pkzip, and 
the predictive PPM family, like PPMZ, are possibly not precisely 
symmetric.
This is related to the stream-based property: the initial file $x$
may have regularities to which the compressor adapts;
after crossing the border to $y$ it must unlearn those regularities
and adapt to the ones of $x$. This process may cause some imprecision
in symmetry that vanishes asymptotically with the length of $x,y$.
A compressor must be poor indeed (and will certainly
not be used to any extent) if it doesn't satisfy symmetry up
to the required precision.
Apart from stream-based, the other major family of compressors 
is block-coding based, like bzip2.
They essentially analyze the 
full input block by considering all rotations in obtaining
the compressed version. It is to a great
extent symmetrical, and real  experiments show no
departure from symmetry.

{\bf Distributivity:}
The distributivity property is not immediately intuitive.
In Kolmogorov complexity theory the stronger distributivity
property 
\begin{equation}\label{eq.sdistr}
C(xyz)+C(z) \leq C(xz)+C(yz)
\end{equation}
holds (with $K=C$). However, to prove
the desired properties of NCD below, only the weaker
distributivity property 
\begin{equation}\label{eq.wdistr}
C(xy)+C(z) \leq C(xz)+C(yz) 
\end{equation}
above is required,
also for the boundary case were $C=K$. In practice, real-world
compressors appear to satisfy this weaker distributivity property up to
the required precision.
%Given the stream-based
%property, the distributivity property follows from the stronger,
%but perhaps intuitively more appealing, \eqref{eq.sap} below. 
%Compressing $xy$, the compressor first compresses $x$ to a file
%$x^*$ of $C(x)$ bits.
%Then, using the regularities extracted from the ``data base'' $x$,
%the compressor compresses $y$ to $(y|x)^*$, 
%to finally halt with a file $(xy)^*$
%of $C(xy)=C(x)+C(y|x)$ bits. 

\begin{definition}
\rm
Define
\begin{equation}\label{eq.cci}
C(y|x)=C(xy)-C(x).
\end{equation}
This number $C(y|x)$ of bits of information in $y$, relative to $x$,
can be viewed as the excess number of bits in
$(xy)^*$ compared to $x^*$,  and is called
the amount of {\em conditional compressed information}.
\end{definition}
In the definition of compressor the decompression algorithm is
not included (unlike the case of Kolmorogov complexity, where
the decompressing algorithm is given by definition), but
it is easy to construct one: 
Given the file $x^*$ in $C(x)$ bits, we can run the compressor on
all candidate strings $z$---for example, in length-increasing lexicographical  
order, until we find the compressed string $z_0=x$. Since this
string decompresses to $x$ we have found $x=z_0$.
Given the file $(xy)^*$ in $C(xy)$ bits
we repeat this process using strings $xz$ until
we find $(xz_1)^*=(xy)^*$. Since
this string decompresses to $xy$, we have found $y=z_1$.
By the unique decompression property we find that $C(y|x)$ is the
extra number of bits we require to describe $y$ apart from 
describing $x$.
It is intuitively acceptable that the conditional compressed information
$C(x|y)$ satisfies the triangle inequality 
\begin{equation}\label{eq.sap}
C(x|y) \leq C(x|z)+C(z|y).
\end{equation}

\begin{lemma}
Both \eqref{eq.sdistr} and \eqref{eq.sap} imply \eqref{eq.wdistr}.
\end{lemma}
\begin{proof}
(\eqref{eq.sdistr} implies \eqref{eq.wdistr}:)
By monotonicity.

(\eqref{eq.sap} implies \eqref{eq.wdistr}:)
Rewrite the terms in \eqref{eq.sap} according to  \eqref{eq.cci},
cancel $C(y)$ in the left- and right-hand sides, use symmetry, and rearrange. 
\end{proof}

\begin{lemma}
A normal compressor satisfies additionally % {\em monotonicity},
%$C(x), C(y) \leq C(xy)$, 
{\em subadditivity}:
$C(xy) \leq C(x)+C(y)$.
\end{lemma}
\begin{proof}
%{\em Monotonicity:}  This follows directly from stream-basedness
%($C(xy) \geq C(x)$), and symmetry together with stream-basedness 
%($C(xy)=C(yx) \geq C(y)$).
%
%{\em Subadditivity:} 
Consider the special case of distributivity 
with $z$ the empty word so that $xz=x$, $yz=y$, and $C(z)=0$.
\end{proof}

{\bf Subadditivity:}
The subadditivity property is clearly also required
for every viable compressor, since a compressor may use information
acquired from $x$ to compress $y$. Minor imprecision may arise from
the unlearning effect of 
crossing the border between $x$ and $y$, 
mentioned in relation to symmetry,
but again this must vanish asymptotically with increasing length of $x,y$.

\subsection{Properties of the NCD:}
There is a natural interpretation to $NCD(x,y)$: If, say, $C(y) \geq C(x)$
then we can rewrite
\[NCD(x,y) = \frac{C(xy)-C(x)}{C(y)} . \]
That is, the distance $NCD(x,y)$ between $x$ and $y$ is the
improvement due to compressing $y$ using $x$ as previously compressed
``data base,'' and compressing $y$ from scratch,
expressed as the ratio between the bit-wise length of the two
compressed versions.
%In analogy with ``mutual information'' between two random variables,
%Kolmogorov complexity theory has a notion of information
%between two individual finite objects. 
Relative to the reference compressor we
can define the information  in $x$ about $y$ as $C(y)-C(y|x)$. Then,
\[
NCD(x,y) = 1 - \frac{C(y)-C(y|x)}{C(y)}.
\]
That is, the NCD between $x$ and $y$ is 1 minus the ratio of the
information $x$ about $y$ and the information in $y$.

\begin{theorem}
If the compressor is normal, then the NCD is a similarity metric.
\end{theorem}

\begin{proof}
({\em Metricity:})
We first show that the NCD satisfies the three metric (in)equalities
up to $O((\log n)/n)$ additive precision, where $n$ is the maximal binary
length of an opject involved in the (in)equality concerned.

(i) By idempotency we have $NCD(x,x)=0$.
By monotonicity we have $NCD (x,y) \geq 0$ for every $x,y$. 

(ii) $NCD(x,y)=NCD(y,x)$. The NCD is unchanged by interchanging
$x$ and $y$ in \eqref{eq.ncd}.

(iii) The difficult property is the triangle inequality.
%By symmetry we have $C(xy)=C(yx)$, and hence we ignore
%the minimization of the first term in the denominator
%of the NCD formula. (For some compressor this may hold only
%up to some precision.) 
Without loss of generality we assume $C(x) \leq C(y) \leq C(z)$.
Since the NCD is symmetrical, there are only three triangle inequalities
that can be expressed by $NCD(x,y), NCD(x,z),NCD(y,z)$.
We verify them in turn:

{\em Proof of} $NCD(x,y) \leq NCD(x,z)+NCD(z,y)$:
By distributivity, the compressor itself 
satisfies 
$C(xy) + C(z) \leq 
C(xz) + C(zy)$. Subtracting $C(x)$ from both sides 
and rewriting, $C(xy) - C(x) \leq
C(xz) - C(x) + C(zy) - C(z)$. 
%\leq C(xz) - C(x) + C(zy) - C(y)$.
Dividing by $C(y)$ on both sides we find
\[ \frac{C(xy) - C(x)}{C(y)} \leq \frac{C(xz) - C(x) + C(zy) - C(z)}{C(y)}.
\]
The left-hand side is $\leq 1$.
                                                                                
{\em Case a:} Assume the right-hand side is $\leq 1$.
Setting $C(z)=C(y)+ \Delta$, and adding $\Delta$ to both the numerator
and denominator of the right-hand side, it can only increase and
draw closer to 1. Therefore,
\begin{align*}
\frac{C(xy)-C(x)}{C(y)} & \leq \frac{C(xz)-C(x) 
+ C(zy)-C(z) + \Delta}{C(y) + \Delta}
\\ & =  \frac{C(zx)-C(x)}{C(z)} + \frac{ C(zy)-C(y)}{C(z)},
\end{align*}
which was what we had to prove.
                                                                                
{\em Case b:} Assume the right-hand side is $>1$.
We proceed like in Case 2.1, and add $\Delta$ to both numerator
and denominator. Although now the right-hand side decreases,
it must still be greater than 1, and therefore the right-hand side
remains at least as large as the left-hand side.

{\em Proof of} $NCD(x,z) \leq NCD(x,y)+NCD(y,z)$:
By distributivity we have $C(xz)+C(y) \leq C(xy)+C(yz)$. Adding
$C(x)$ to both sides, and dividing both sides by $C(z)$ we obtain
\[
\frac{C(xz)-C(x)}{C(z)} \leq \frac{C(xy)-C(x)}{C(z)}
+\frac{C(yz)-C(y)}{C(z)}.
\]
The right-hand side doesn't decrease when we substitute $C(y)$ for the
denominator $C(z)$ of the first term, since $C(y) \leq C(z)$.
Therefore, the inequality stays valid under this substitution, which was
what we had to prove.

{\em Proof of} $NCD(y,z) \leq NCD(y,x)+NCD(x,z)$:
By distributivity we have $C(yz)+C(x) \leq C(yx)+C(xz)$. Adding
$C(y)$ to both sides, and dividing both sides by $C(z)$ we obtain
\[
\frac{C(yz)-C(y)}{C(z)} \leq \frac{C(yx)-C(x)}{C(z)}
+\frac{C(yz)-C(y)}{C(z)}.
\]
The right-hand side doesn't decrease when we substitute $C(y)$ for the
denominator $C(z)$ of the first term, since $C(y) \leq C(z)$.
Therefore, the inequality stays valid under this substitution, which was
what we had to prove.

({\em Range:}) We next verify the range of the NCD:
The $NCD(x,y)$ is always in between 0 and 1,
up to an additive $O((\log n)/n)$  term with
$n$ the maximum binary length of $x,y$.

({\em Density:})
It remains to show that the NCD satisfies \eqref{eq.dp}.
Since $C(y) \leq C(x)$, the set condition $NCD(x,y) \leq e$
can be rewritten (using symmetry) as
$C(xy)-C(y) \leq eC(x)$. Using $C(y) \leq C(x)$ again, we also have
$C(xy)-C(x) \leq  C(xy)-C(y)$. Together, 
\begin{equation}\label{eq.dens}
C(xy) \leq (1+e)C(x). 
\end{equation}
Suppose that for some $e$ there
are $\geq 2^{(1+e)C(x)+1}$ distinct $xy$ satisfying
\eqref{eq.dens}. Since there are at most
$\sum_{i=0}^{(1+e)C(x)} 2^i = 2^{(1+e)C(x)+1}-1$ programs to encode the
concatenations $xy$,  by the pigeon hole principle,
two distinct ones share the same code word.
This contradicts the unique decompression
property.
\end{proof}

{\bf Quasi-Universality:}
We now digress to the theory developed in \cite{Li03}, which
formed the motivation for developing the NCD.
If, instead of the result of some real compressor,
we substitute the Kolmogorov
complexity for the lengths of the compressed files in the NCD
formula, the result is a similarity
metric. Let us call it the {\em Kolmogorov metric}.
It is universal
in the following sense: Every metric expressing similarity 
according to some feature,
that can be computed from the objects concerned, is comprised
(in the sense of minorized) by the universal metric. 
Note that every feature of the data gives rise to a similarity,
and, conversely, every similarity can be thought of as expressing some feature:
being similar in that sense. 
Our actual practice in using the NCD falls short of 
this ideal theory in at least 
three respects:

(i) The claimed universality of the Kolmogorov metric
holds only for indefinitely long sequences $x,y$. Once we consider
strings $x,y$ of definite length $n$, the Kolmogorov metric
is only universal with respect to ``simple'' computable normalized information
distances, where ``simple'' means that they are computable by programs
of length, say, logarithmic in $n$.
This reflects the fact that, technically speaking, the universality 
is achieved by summing the weighted contribution of all
similarity metrics in the class considered with respect
to the objects considered. Only similarity metrics of which
the complexity is small (which means that the weight is large)
with respect to the size of the data concerned kick in. 

(ii) The Kolmogorov complexity is not computable, and it is 
in principle impossible to compute how far off the
NCD is from the Kolmogorov metric. So we cannot in general know
how well we are doing using the NCD.

(iii) To approximate the NCD
we use standard compression programs like gzip, PPMZ, and bzip2. 
While better compression
of a string will always  approximate the Kolmogorov complexity better,
this may not be true for the NCD. Due to its arithmetic
form, subtraction and division, it is theoretically possible
that while all items in the formula get better compressed, 
the improvement is not the same for all items, and the NCD value
moves away from the asymptotic value.  
%There, we consider the difference of
%two compressed quantities in the numerator.
%Different compressors may compress
%the two quantities differently, causing an increase in the
%difference even when both quantities are compressed better (but
%not both as well). We
%also have to deal with a ratio that causes a similar problem.
%Thus, a better compression program may not necessarily mean
%that we also approximate the (normalized) information distance
%better. This was borne out by the results of our experiments using
%different compressors. 
%Our particular form of expressing the NCD
%with $\min$ terms in the denominator tries to compensate for this
%(Remark~\ref{rem.compensate}). 
In our experiments we have not observed this behavior in a noticable fashion.
Formally, we can state the following:

\begin{theorem}
Let $d$ be a computable similarity metric.
Given a constant $a \geq 0$, let objects $x,y$
be such that $C(xy)-K(xy) \leq a$. 
Then, $NCD(x,y) \leq d(x,y) + (a+O(1))/k$ where $k=\max\{C(x),C(y)\}$. 
\end{theorem}

\begin{proof}
Fix $d,C,x,y,k$ in the statement of the theorem.
Without loss of generality we assume $d(x,y)=e$ and $C(x)=k$.
By \eqref{eq.dp}, there are  
$< 2^{(1+e)k+1}$ many
$(u,v)$ pairs,
such that $C(u) \leq C(v)=k$ and $d(u,v) \leq e$. 
Since both $C$ and $d$ are computable
functions, we can compute and enumerate all these pairs $(u,v)$.
The initially fixed pair $(x,y)$ is an element in the list and its
index takes $\leq (1+e)k+1$ bits. 
It can be described by at most $(1+e)k+O(1)$ bits---its index in the list 
and an $O(1)$ term accounting for the lengths of the programs involved
in reconstructing $(x,y)$ given its index in the list, including
algorithms to compute functions $d$ and $C$.
Since the Kolmogorov complexity gives the length of the shortest
effective description, we have $K(xy) \leq (1+e)k+O(1)$.
Then, 
\[
NCD(x,y) = \frac{C(xy)-C(x)}{C(x)} \leq \frac{K(xy)-C(x) +a}{C(x)}
\leq \frac{eC(x)+a+O(1)}{C(x)}  \leq e + \frac{a+O(1)}{C(x)}.
\]
\end{proof}

So the NCD is {\em quasi-universal} in the sense that if for objects
$x,y$ the 
compression complexity $C(xy)$ approximates the 
Kolmogorov complexity $K(xy)$ up to much closer precision
than $C(x)$, then the $NCD(x,y)$ minorizes the metric $d(x,y)$
up to a vanishing term. More precisely, the error term
$(a+O(1))/C(x) \rightarrow 0$ for a sequence of pairs $x,y$ 
with $C(y) \leq C(x)$, $C(xy)-K(xy) \leq a$ and $C(x) \rightarrow \infty$.

\begin{remark}
\rm
Clustering according to NCD will group sequences together that
are similar according to features that are not 
explicitly known to us. Analysis
of what the compressor actually does, still  may not tell us which 
features that make sense to us can be expressed by conglomerates
of features analyzed by the compressor. This can be exploited to
track down unknown features implicitly: forming automatically 
clusters of data and see in which cluster (if any) a new candidate
is placed.

Another aspect that can be exploited is exploratory:
Given that the NCD is small for a pair $x,y$ of specific sequences,
what does this really say about the sense in which these two sequences are 
similar? 
The above analysis suggests that close
similarity will be due to a dominating feature (that perhaps expresses
a conglomerate of subfeatures). Looking into these deeper causes may give
feedback about the appropriateness of the realized NCD distances and may help
extract more intrinsic information about subject matter
than the oblivious division into clusters by
looking for the common features in the data clusters.
\end{remark}

\section{Clustering}

Given a set of objects, the pairwise NCD's form
the entries of a distance matrix.
This distance matrix contains the pairwise relations 
in raw form. But in this format
that information is not easily usable.
Just as the distance matrix is a reduced form of information
representing the original data set, we now need to reduce the
information even further in order to achieve a cognitively acceptable
format like data clusters.
To extract a hierarchy of clusters
from the distance matrix,
we determine a dendrogram (binary tree) that agrees
with the distance matrix according to a cost measure.
This allows us to extract more information from the data
than just flat clustering (determining disjoint
clusters in dimensional representation).

Clusters are groups of objects that are similar
according to our metric. There are various ways
to cluster. Our aim
is to analyze data sets for which the number of clusters is
not known a priori, and the data are not labeled. As stated in \cite{DHS},
conceptually simple, hierarchical clustering is among
the best known unsupervised methods in this setting, and
the most natural way is to represent the relations
in the form of a dendrogram, which is customarily a directed binary tree
or undirected ternary tree.
To construct the tree from a distance matrix with entries
consisting of the pairwise distances between objects,
we use a quartet method. This is a matter of choice only, other methods
may work equally well.
The distances we compute in our experiments are often
within the range 0.85 to 1.2. That is, the distinguishing
features are small, and we need a sensitive method
to extract as much information contained in the distance matrix as
is possible.  
For example, our experiments showed
that reconstructing a minimum spanning tree is not sensitive
enough and gives poor results.
With increasing number of data items, the projection of the NCD matrix
information into the tree representation format gets increasingly distorted.
A similar situation arises in using alignment cost in genomic comparisons.
Experience shows that in both cases the hierarchical clustering methods 
seem to work best for small sets of data, up to 25 items, and to deteriorate
for larger sets, say 40 items or more. A standard solution to hierarchically
cluster larger sets of data is to first cluster nonhierarchically,
by say multidimensional scaling of $k$-means, available in standard packages,
for instance {\em Matlab},
and then apply hierarchical clustering on the emerging clusters.

\begin{figure}[htb]
\begin{center}
\epsfig{file=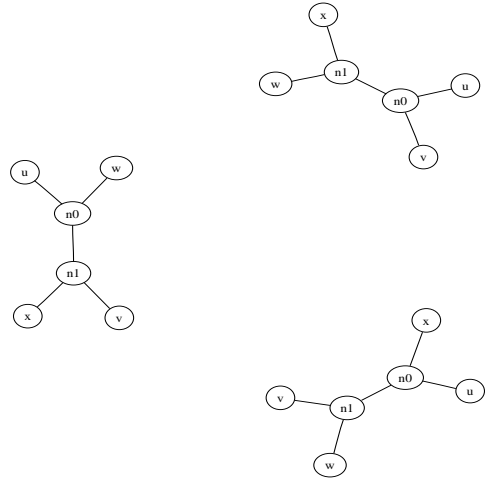,width=2.5in,height=2.5in}
\end{center}
\caption{The three possible quartet topologies for the set of leaf labels {\em u,v,w,x} }\label{figquart}
\end{figure}

{\bf The quartet method:} 
We consider every group of four elements from our set
of $n$ elements;
there are ${n \choose 4}$ such groups.
From each group $u,v,w,x$ we construct a tree of arity 3,
which implies that the tree consists of two subtrees of two
leaves each. Let us call such a tree a {\em quartet topology}.  There are
three possibilities denoted (i) $uv | wx$, (ii) $uw | vx$,
and (iii)  $ux | vw$, where a vertical bar divides the two pairs of leaf nodes
into two disjoint subtrees (Figure~\ref{figquart}).

For any given tree $T$ and any group
of four leaf labels $u,v,w,x$, we say $T$ is $consistent$ with $uv | wx$
if and only if the path from $u$ to $v$ does not cross
the path from $w$ to $x$.  Note that exactly one of the three possible
quartet topologies for any set of 4 labels must be consistent for any given tree.
\begin{figure}[htb]
\begin{center}
\epsfig{file=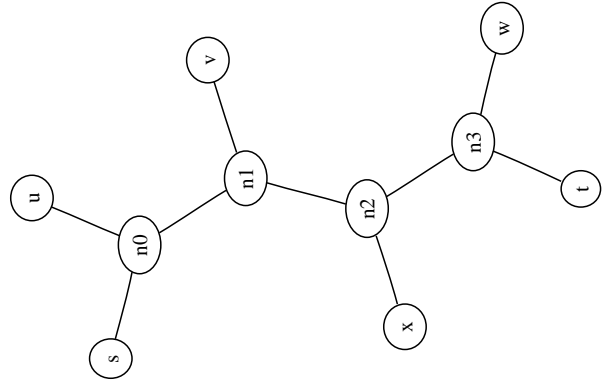,width=2in,angle=90}
\end{center}
\caption{An example tree consistent with quartet topology $uv | wx$ }\label{figquartex}
\end{figure}
We may think of a large tree having many smaller quartet topolgies embedded
within its structure. Commonly the goal in the quartet method
is to find (or approximate as closely as possible) the tree
that embeds the maximal number of consistent (possibly weighted) quartet
topologies from a given set $Q$ of quartet topologies
\cite{Ji01} (Figure~\ref{figquartex}).
This is called the (weighted) {\em Maximum Quartet Consistency (MQC)}
problem.

We propose a new optimization problem: the 
{\em Minimum Quartet Tree Cost (MQTC)}, as follows: 
The cost of a quartet topology is defined as the sum 
of the distances between each pair of neighbors; that
is, $C_{uv|wx} = d(u,v) + d(w,x)$.  
The total cost $C_T$ of a tree $T$ with a set $N$ of leaves (external nodes
of degree 1) is defined as 
$C_T =\sum_{\{u,v,w,x\} \subseteq N} \{C_{uv|wx}: T$ is consistent
with  $uv |wx\}$---the
sum of the costs of all its consistent quartet topologies. 
First, we generate a list of all possible quartet topolgies 
for all four-tuples of labels
under consideration.  For each group of 
three possible quartet topologies for a given
set of four labels $u,v,w,x$, calculate a best (minimal) cost
$m(u,v,w,x) = \min \{ C_{uv|wx}, C_{uw|vx}, C_{ux|vw} \}$, 
and a worst (maximal)
cost $M(u,v,w,x) = \max \{ C_{uv|wx}, C_{uw|vx}, C_{ux|vw} \}$.  
Summing all best quartet toplogies yields the best (minimal) cost 
$m = \sum_{\{u,v,w,x\} \subseteq N} m(u,v,w,x)$.
Conversely, summing all worst quartet toplogies yields the worst (maximal) cost
$M =  \sum_{\{u,v,w,x\} \subseteq N} M(u,v,w,x)$.
For some distance matrices,
these minimal and maximal values can not be attained by actual trees;
however, the score $C_T$ of every tree $T$ will lie between these two values.
In order to be able to compare tree scores in a more uniform way,
we now rescale the score linearly such that the worst score maps to 0,
and the best score maps to 1, and term this the 
{\em normalized tree benefit score} $S(T) = (M-C_T)/(M-m)$.
Our goal is to find a full tree with a maximum value
of $S(T)$, which is to say, the lowest total cost.

To express the notion of computational difficulty one uses
the notion of ``nondeterministic polynomial time (NP)''.
If a problem concerning $n$ objects is NP-hard 
this means that the best known algorithm
for this (and a wide class of significant problems) requires
computation time exponential in $n$. That is, it is infeasible
in practice. 
The {\em MQC decision problem} is the following:
Given $n$ objects, let $T$ be a tree of which the $n$
leaves are labeled by the objects, and let 
$Q_T$ be the set of quartet topologies 
embedded in $T$.
Given a set of quartet topologies $Q$,
and an integer $k$,
the problem is to decide whether there is
a binary tree $T$ such that $Q \bigcap Q_T > k$. 
In \cite{Ji01} it is shown that the MQC decision problem 
is NP-hard.
   For every MQC decision problem one can define an
MQTC problem that has the same solution: give the 
quartet topologies in $Q$ cost 0 and the other ones cost 1.
This way the MQC decision problem can be
reduced to the MQTC decision problem, which shows also the latter
to be NP-hard. Hence,
it is infeasible in practice, but we can sometimes solve it, and
always approximate it. 
(The reduction also shows that the quartet problems reviewed in
\cite{Ji01}, 
are subsumed by our problem.)  
Adapting current
methods in \cite{Br00} to our MQTC optimization problem,
results in far too computationally 
intensive calculations;
they run many months or years on moderate-sized problems
of 30 objects. Therefore, we have designed a 
simple, feasible, heuristic method for our problem based
on randomization and hill-climbing.  First, a random tree with $2n-2$ nodes
is created, consisting of $n$ leaf nodes (with 1 connecting edge) labeled 
with the names of the data items, and $n-2$ non-leaf or {\em internal} nodes
labeled with the lowercase letter ``n'' followed by a unique integer identifier.  Each internal node has exactly three connecting edges.  For this 
tree $T$, we calculate the total cost of all embedded quartet toplogies, 
and invert and scale this value to 
find $S(T)$.  A tree is consistent with precisely
$\frac{1}{3}$ of all quartet topologies, one for every quartet.
A random tree may be consistent with about $\frac{1}{3}$ of the best
quartet topologies---but because of dependencies this figure is
not precise. The initial random
this tree is chosen as the currently best known tree, and is used as
the basis for further searching.  We define a simple mutation on a tree
as one of the three possible transformations:
\begin{enumerate}
\item A {\em leaf swap}, which consists of randomly choosing two leaf nodes
and swapping them.
\item A {\em subtree swap}, which consists of randomly choosing two internal 
nodes and swapping the subtrees rooted at those nodes.
\item A {\em subtree transfer}, whereby a randomly chosen subtree (possibly a leaf) is detached and reattached in another place, maintaining arity invariants.
\end{enumerate}
Each of these simple mutations keeps the
number of leaf nodes and internal nodes in the tree invariant; 
only the structure and placements
change.  Define a full mutation as a sequence of at least one but potentially
many simple mutations, picked according to the following distribution. 
First we pick the number $k$ of simple mutations that we will perform with
probability $2^{-k}$.  For each such simple mutation, we choose one of
the three types listed above with equal probability.  Finally, for each of 
these simple mutations, we pick leaves or internal nodes, as necessary.  Notice
that trees which are close to the original tree (in terms of number of 
simple mutation steps in between) are examined often, while trees that are 
far away from the original tree will eventually be examined, but not very 
frequently.
In order to search for a better tree,
we simply apply a full mutation on $T$ to arrive at $T'$, and then
calculate $S(T')$.  If $S(T') > S(T)$, then keep $T'$ as the new best tree.
Otherwise, try a new different tree and repeat.  If $S(T')$ ever reaches
$1$, then halt, outputting the best tree.  Otherwise, run until it seems
no better trees are being found in a reasonable amount of time, in which
case the approximation is complete.

\begin{figure}[htb]
\begin{center}
\epsfig{file=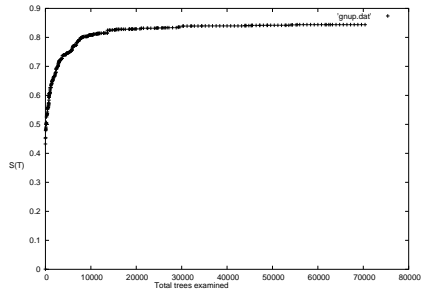,width=1.5in,angle=270}
\end{center}
\caption{Progress of a 60-item data set experiment over time}\label{figprogress}
\end{figure}

Note that if a tree is ever found such that $S(T) = 1$, then we can stop
because we can be certain that this tree is optimal, as no tree could 
have a lower cost.  In fact, this perfect tree result is achieved in our 
artificial tree reconstruction experiment (Section~\ref{sect.artificial}) 
reliably in a few minutes.  For real-world data, $S(T)$ reaches 
a maximum somewhat 
less than $1$, presumably reflecting distortion of the information
in the distance matrix 
data by the best possible tree representation, as noted above,
or indicating getting stuck in a local optimum or a search space too large
to find the global optimum.
On many typical problems of up to 40 objects this tree-search gives a tree 
with $S(T) \geq 0.9$ within half an hour.  For large numbers of objects,
tree scoring itself can be slow (as this takes order $n^4$ computation steps), 
and the space of
trees is also large, so the algorithm may slow down substantially.
For larger experiments, we use a C++/Ruby implementation with MPI (Message
Passing Interface, a common standard used on massively parallel computers) on a
cluster of workstations in parallel to find trees more rapidly. We can
consider the graph 
mapping the achieved $S(T)$ score as a function
of the number of trees examined.  Progress
occurs typically in a sigmoidal fashion towards a maximal value $\leq 1$,
Figure~\ref{figprogress}.

\subsection{Three controlled experiments}\label{sect.artificial}

With the natural data sets we use, one may have the preconception 
(or prejudice) that, say,  music by Bach should be clustered together, 
music by Chopin should be clustered together, and so should music by
rock stars. However, the preprocessed music files of a piece by Bach and
a piece by Chopin, or the Beatles, may resemble one another 
more than two different
pieces by Bach---by accident or indeed by design and copying. Thus, natural
data sets may have ambiguous, conflicting, or counterintuitive 
outcomes. In other words, the experiments on natural data sets have
the drawback of not having an objective clear ``correct'' answer that can 
function as a benchmark for assessing our experimental outcomes,
but only intuitive or traditional preconceptions.
We discuss three experiments that show that our
program indeed does what it is supposed to do---at least in 
artificial situations where we know in advance what the correct answer is.
The similarity machine consists of two parts: (i) extracting a distance matrix
from the data, and (ii) constructing a tree 
from the distance matrix using our novel quartet-based heuristic.

\begin{figure}[htb]
\begin{center}
\epsfig{file=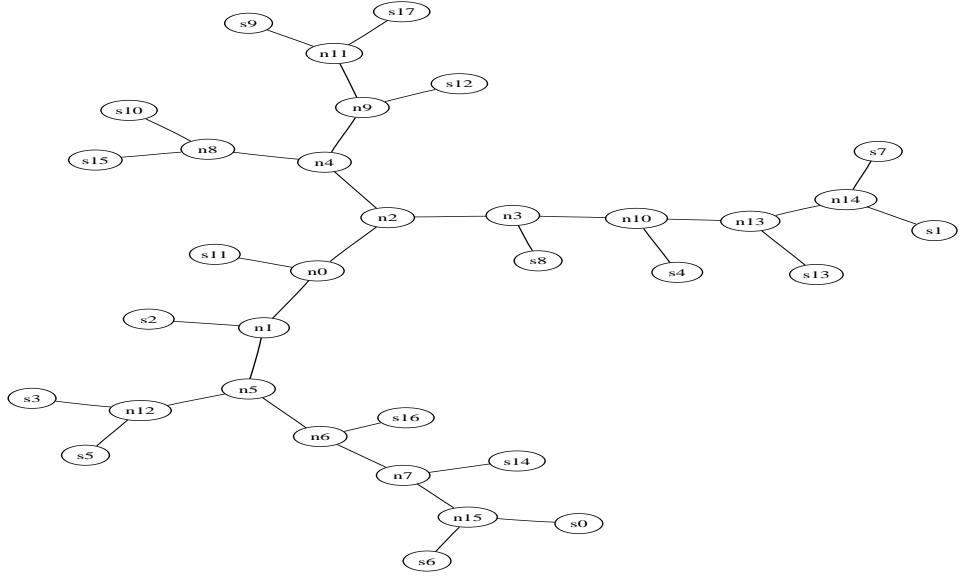,width=5in,height=3in}
\end{center}
\caption{The randomly generated tree that our algorithm reconstructed. $S(T)=1$.}\label{figarttreereal}
\end{figure}

{\bf Testing the quartet-based tree construction:}
We first test whether the quartet-based tree construction
heuristic is trustworthy: 
We generated a ternary tree $T$ with 18 leaves, using the pseudo-random
number generator ``rand'' of the Ruby programming language,
and derived
a metric from it by defining the distance between 
two nodes as follows:
Given the length of the path from $a$ to $b$, in an integer number of
edges, as $L(a,b)$, let 
\[d(a,b) = { {L(a,b)+1} \over 18},
\]
  except when
$a = b$, in which case $d(a,b) = 0$.  It is easy to verify that this
simple formula always gives a number between 0 and 1, and is monotonic
with path length.
Given only the $18\times 18$ matrix of these normalized distances, 
our quartet method exactly reconstructed  the original tree
$T$ represented in
Figure~\ref{figarttreereal}, 
with $S(T)=1$.
%TODO: Rudi, Paul wants the distance matrix included here as well

\begin{figure}[htb]
\begin{center}
\epsfig{file=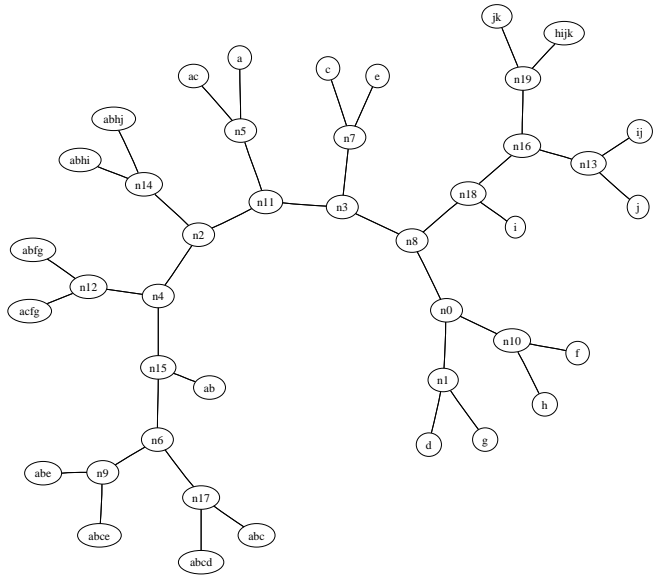,height=3in}
\end{center}
\caption{Classification of artificial files with repeated 1-kilobyte tags. 
 Not all possiblities
are included; for example, file ``$b$'' is missing.
$S(T)=0.905$. }\label{figtaggedfiles}
\end{figure}
\begin{figure}[htb]
\begin{center}
\epsfig{file=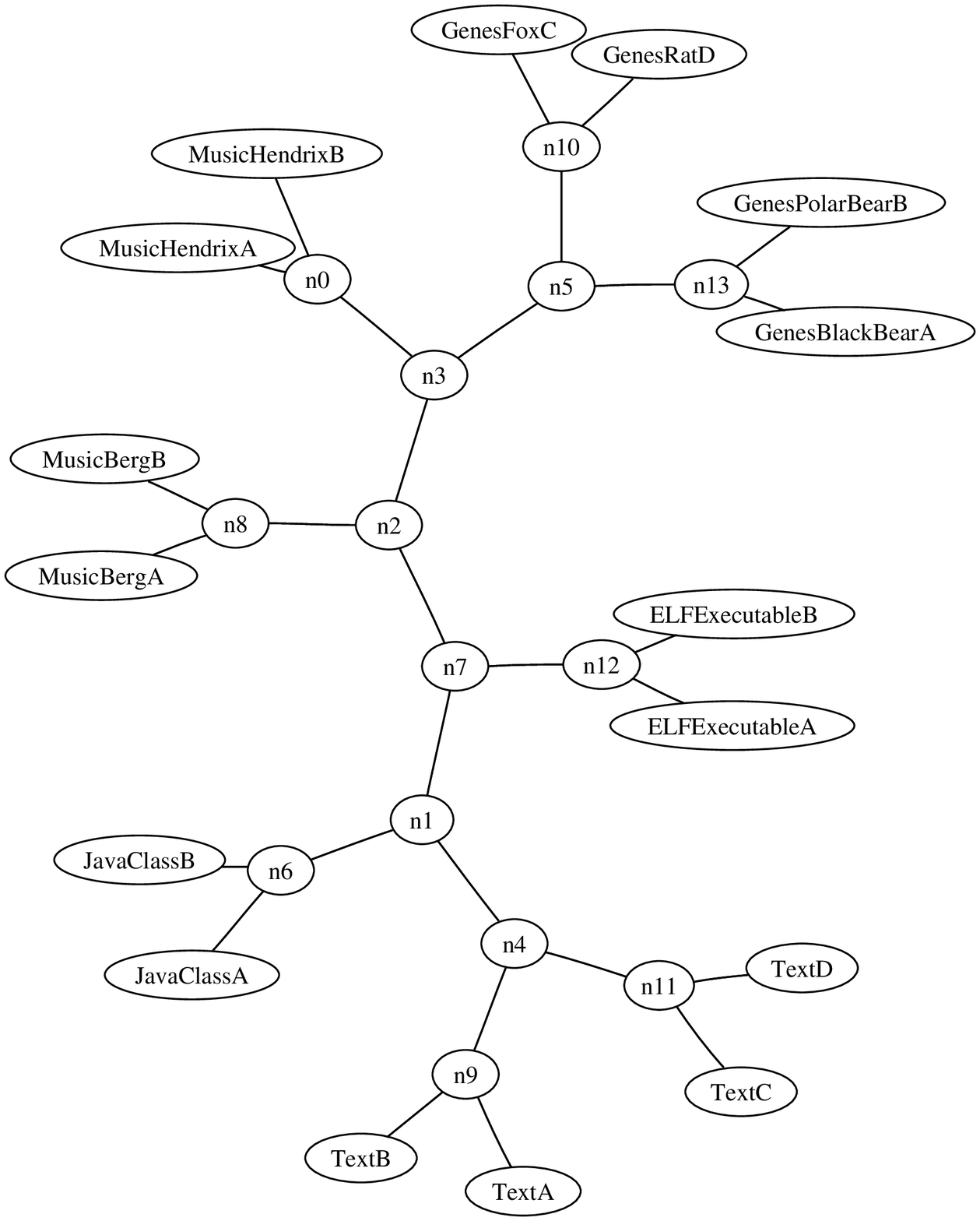,height=4in}
\end{center}
\caption{Classification of different file types. 
Tree agrees 
exceptionally well with
NCD distance matrix: $S(T)=0.984$.}\label{figfiletypes}
\end{figure}

{\bf Testing the similarity machine on artificial data:}
Given that the tree reconstruction method is accurate
on clean consistent data, we tried whether the full procedure
works in an acceptable manner when we know what the outcome should
be like. We used the ``rand'' pseudo-random number generator
from the C programming language standard library under Linux.
We randomly generated 11 separate 1-kilobyte blocks of data where
each byte was equally probable and called these {\em tags}.  Each tag
was associated with a different lowercase letter of the alphabet.  Next,
we generated 22 files of 80 kilobyte each, 
by starting with a block of purely random
bytes and applying one, two, three, or four different tags on it.
Applying a tag consists of ten repetitions of picking a random location
in the 80-kilobyte file, and overwriting that location with the globally
consistent tag that is indicated.  So, for instance, to create the file
referred to in the diagram by ``a,'' we start with 80 kilobytes of random data,
then pick ten places to copy over this random data with the arbitrary 
1-kilobyte sequence identified as tag {\em a}.  
Similarly, to create file ``ab,''
we start with 80 kilobytes of random data, then pick ten places to put
copies of tag {\em a}, then pick ten more places to put copies of tag {\em b} (perhaps
overwriting some of the {\em a} tags).  Because we never use more than four
different tags, and therefore never place more than 40 copies of tags, we
can expect that at least half of the data in each file is random and
uncorrelated with the rest of the files.  The rest of the file is 
correlated with other files that also contain tags in common; the more 
tags in common, the more related the files are.
The compressor used to compute the NCD matrix was bzip2.
The resulting tree 
 is given in Figure~\ref{figtaggedfiles}; it can be
seen that 
the clustering has  occured exactly as we would expect.
The $S(T)$ score is 0.905.

{\bf Testing the similarity machine on natural data:}
We test gross classification of files
based on markedly different file types.  Here, we chose several files:
(i) Four mitochondrial gene sequences, from a black bear, polar bear, 
fox, and rat obtained from the GenBank Database on the world-wide web;
(ii) Four excerpts from the novel { \em The Zeppelin's Passenger} by 
E.~Phillips Oppenheim, obtained from the Project Gutenberg Edition
on the World-Wide web;
(iii) Four MIDI files without further processing; two from Jimi Hendrix and 
two movements from Debussy's Suite Bergamasque, downloaded from various
repositories on the
world-wide web;
(iv) Two Linux x86 ELF executables (the {\em cp} and {\em rm} commands),
copied directly from the RedHat 9.0 Linux distribution; and
(v)  Two compiled Java class files, generated by ourselves.
The compressor used to compute the NCD matrix was bzip2.
As expected, the program correctly classifies each of the different types
of files together with like near like. The result is reported
in Figure~\ref{figfiletypes} with $S(T)$ equal to the very high
confidence value 0.984.
This experiment shows the power and universality of the method:
no features of any specific domain of application are used.

\section{Experimental Validation}
We developed the CompLearn Toolkit, Section~\ref{sect.intro}, and performed
experiments in vastly different
application fields to test the quality and universality of the method. 
The success of the method as reported below depends strongly on the
judicious use of encoding of the objects compared. Here one should
use common sense on what a real world compressor can do. There are 
situations where our approach fails if applied in a
straightforward way.
For example: comparing text files by the same authors
in different encodings (say, Unicode and 8-bit version) is bound to fail.
For the ideal similarity metric  based on
Kolmogorov complexity as defined in \cite{Li03}
this does not matter at all, but for
practical compressors used in the experiments it will be fatal.
Similarly, in the music experiments below we use symbolic MIDI
music file  format rather than wave format music files. The reason is that
the strings resulting from straightforward 
discretizing the wave form files may be too sensitive to how we discretize.

\subsection{Genomics and Phylogeny}
\begin{figure}
\hfill\ \psfig{figure=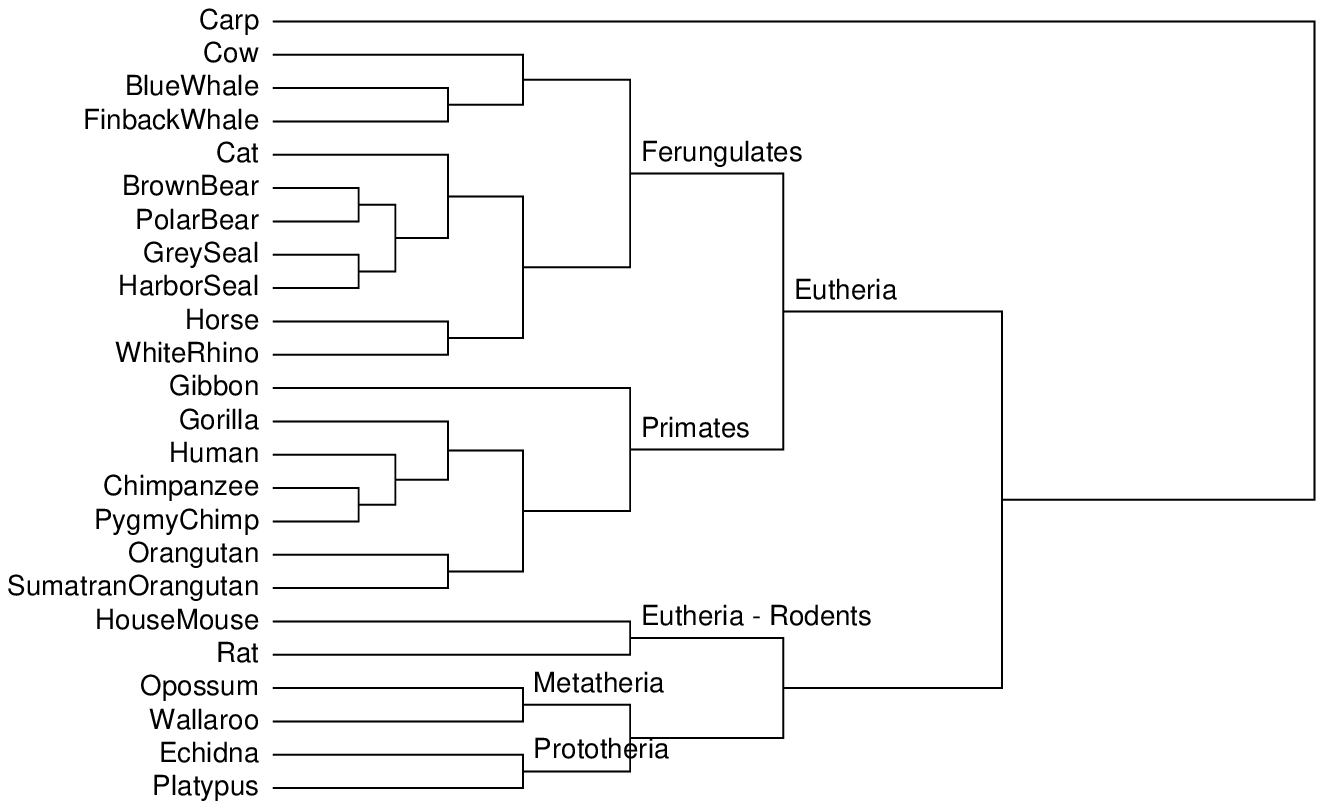,width=3.5in,height=2in} \hfill\
\caption{The evolutionary tree built from complete mammalian mtDNA
sequences of 24 species, using the NCD matrix of Figure~\ref{fig.distmatr}.
We have redrawn the tree from our output to agree
better with the customary phylogeny tree format. The tree agrees exceptionally
well with the NCD distance matrix: $S(T)=0.996$.
}
\label{tree-mammal}
\end{figure}
\begin{figure}
\hfill\ \psfig{figure=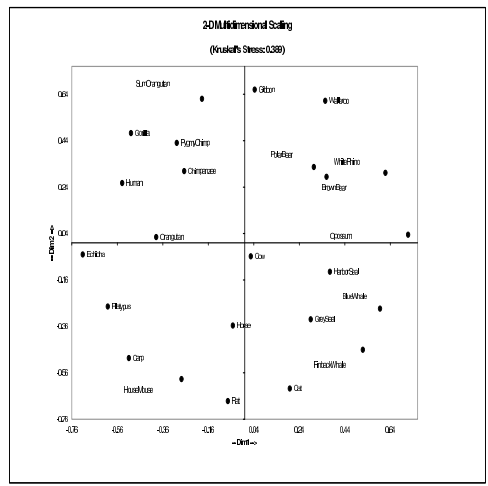,width=2.5in,height=2.5in,angle=270} \hfill\
\caption{Multidimensional clustering of same NCD matrix 
(Figure~\ref{fig.distmatr}) as used for Figure~\ref{tree-mammal}.
Kruskal's stress-1 = 0.389.
}\label{fig.mammal2d}
\end{figure}
In recent
years, as the complete genomes of various species become available,
it has become possible to do whole genome phylogeny (this overcomes the
problem that useing different targeted parts
of the genome, or proteins, may give different trees \cite{RWKC03}). 
Traditional phylogenetic methods on individual genes depended
on multiple alignment of the related proteins
and on the model of evolution of individual amino acids.
Neither of these is practically applicable to the genome level.
In absence of such models, a method which can compute the shared
information between two sequences is useful because biological
sequences encode information, and the occurrence of evolutionary
events (such as insertions, deletions, point mutations,
rearrangements, and inversions) separating two sequences sharing a
common ancestor will result in the loss of their shared
information. Our method (in the form of the CompLearn Toolkit) is
a fully automated 
software tool based on such a distance to compare two
genomes. 
\paragraph{Mammalian Evolution:} In evolutionary biology the 
timing and origin of the major extant placental clades
(groups of organisms that have evolved from a common ancestor)
continues to fuel debate and research. Here, we provide evidence
by whole mitochondrial genome phylogeny for competing hypotheses
in two main questions: the grouping of the Eutherian orders,
and the Therian hypothesis versus the Marsupionta hypothesis. 
\subparagraph{Eutherian Orders:}
We demonstrate (already in \cite{Li03}) that
a whole mitochondrial genome phylogeny of the Eutherians (placental
mammals) 
can be reconstructed automatically from {\em
unaligned} complete mitochondrial genomes by use of an
early form of our compression method,
using standard software packages.
%We will use whole mitochondrial DNA genomes of 20 mammals and the
%problem of Eutherian orders to
%make a comprehensive examination of our measures.
%The problem we consider is this:
As more genomic material has become available,
the debate in biology has intensified
 concerning which two of the three main groups of placental
mammals are more closely related: Primates, Ferungulates, and Rodents.
In \cite{Cao1998}, the maximum likelihood method of phylogeny tree
reconstruction gave evidence for the
(Ferungulates, (Primates, Rodents)) grouping for half of
the proteins in the mitochondial genomes investigated, and
(Rodents, (Ferungulates, Primates)) for the other halves of the mt genomes. 
In that experiment
they aligned 12 concatenated
mitochondrial proteins, taken from 20 species: rat ({\em Rattus
norvegicus}), house mouse ({\em Mus musculus}), 
grey seal ({\em
Halichoerus grypus}), harbor seal ({\em Phoca vitulina}), cat ({\em
Felis catus}), white rhino ({\em Ceratotherium simum}), horse ({\em
Equus caballus}), finback whale ({\em Balaenoptera physalus}), blue
whale ({\em Balaenoptera musculus}), cow ({\em Bos taurus}), gibbon
({\em Hylobates lar}), gorilla ({\em Gorilla gorilla}), human ({\em
Homo sapiens}), chimpanzee ({\em Pan troglodytes}), pygmy chimpanzee
({\em Pan paniscus}), orangutan ({\em Pongo pygmaeus}), Sumatran
orangutan ({\em Pongo pygmaeus abelii}), using opossum ({\em Didelphis
virginiana}), wallaroo ({\em Macropus robustus}), and the platypus ({\em
Ornithorhynchus anatinus}) as outgroup.
In \cite{Li01,Li03}
we used the whole mitochondrial genome of the same
20 species, computing the NCD distances 
(or a closely related distance in \cite{Li01}),
using the GenCompress compressor, followed
by tree reconstruction using the neighbor joining program
in the MOLPHY package \cite{SN87} to confirm the 
commonly believed morphology-supported hypothesis 
(Rodents, (Primates, Ferungulates)). Repeating the experiment using
the hypercleaning method \cite{Br00} of phylogeny tree reconstruction 
gave the same result. Here,
we repeated this experiment several times using the CompLearn Toolkit 
using our new quartet method for reconstructing trees, and
computing the NCD
with various compressors (gzip, bzip2, PPMZ), 
again always with the same result. These experiments are not reported since they
are subsumed by the larger experiment of Figure~\ref{tree-mammal}.
\subparagraph{Marsupionta and Theria:}
The extant monophyletic divisions of the class Mammalia are
the Prototheria (monotremes: mammals that procreate using eggs), 
Metatheria (marsupials: mammals that procreate using pouches), and 
Eutheria (placental mammals: mammals that procreate using placentas). 
The sister relationships between
these groups is viewed as the most fundamental question in
mammalian evolution  \cite{KBSMJ01}. Phylogenetic comparison by either
anatomy or mitochondrial genome has resulted in two conflicting
hypotheses: the gene-isolation-supported {\em Marsupionta hypothesis}: 
((Prototheria, Metatheria), Eutheria) 
versus the morphology-supported {\em Theria hypothesis}:  
(Prototheria, (Methateria, Eutheria)),
the third possiblity apparently not being held seriously by anyone.
There has been a lot of support for either hypothesis;
recent support for the Theria hypothesis was given
in \cite{KBSMJ01} by analyzing a large nuclear gene (M6P/IG2R),
viewed as important
across the species concerned, and even more recent support for
the Marsupionta hypothesis was given in \cite{JMWWA02} by
phylogenetic analysis of another sequence from the nuclear gene
(18S rRNA) and by the whole mitochondrial genome.   
\subparagraph{Experimental Evidence:} To test the Eutherian orders simultaneously
with the Marsupionta- versus Theria hypothesis, 
we added four animals to the above twenty: 
Australian
echidna ({\em Tachyglossus aculeatus}), 
brown bear ({\em Ursus arctos}),
polar bear ({\em Ursus maritimus}), 
using the common carp 
({\em Cyprinus carpio}) as the outgroup.
Interestingly, while there are many species of Eutheria and Metatheria,
there are only three species of now living Prototheria known: platypus,
and two types of echidna (or spiny anteater). So our sample of the Prototheria
is large.
The addition of the new species might be risky in that
the addition of new relations is known to distort the 
previous phylogeny in traditional computational genomics practice.
With our method, using the full genome and obtaining a single tree with
a very high confidence $S(T)$ value, that risk is not as great as in traditional
methods obtaining ambiguous trees with bootstrap (statistic support) 
values on the edges. 
The mitochondrial genomes of the total of 24 species we used  were 
downloaded from the
GenBank Database on the world-wide web. Each is around 17,000 bases.
The NCD distance matrix was computed using the compressor PPMZ.
The resulting phylogeny, with an almost maximal $S(T)$ score
of 0.996 supports anew the currently
accepted grouping (Rodents, (Primates, Ferungulates)) of the Eutherian orders,
and additionally the Marsupionta hypothesis 
((Prototheria, Metatheria), Eutheria),  
see Figure \ref{tree-mammal}. Overall, our whole-mitochondrial
NCD analysis supports the following hypothesis:
\[
\rm
\overbrace{(\underbrace{(primates, ferungulates)(rodents}_{\rm Eutheria},
(Metatheria, Prototheria)))}^{Mammalia},
\]
which indicates that the rodents, and the branch leading to 
the Metatheria and Prototheria, split off early from the branch that
led to the primates and ferungulates. Inspection of the distance matrix shows
that the primates are very close together, as are the rodents, the Metatheria,
and the Prototheria. These are tightly-knit groups with relatively close
NCD's. The ferungulates are a much looser group with generally distant
NCD's. The intergroup distances show that the Prototheria are
furthest away from the other groups, followed by the Metatheria and
the rodents. 
Also the fine-structure of the tree is consistent with biological wisdom.
\begin{center}
\begin{figure}
{\tiny
\begin{verbatim}
             BlueWhale            Cat             Echidna           Gorilla            Horse            Opossum          PolarBear         SumOrang
                   BrownBear         Chimpanzee         FinWhale          GreySeal        HouseMouse         Orangutan         PygmyChimp         Wallaroo
                           Carp               Cow              Gibbon          HarborSeal          Human            Platypus            Rat            WhiteRhino
     BlueWhale 0.005 0.906 0.943 0.897 0.925 0.883 0.936 0.616 0.928 0.931 0.901 0.898 0.896 0.926 0.920 0.936 0.928 0.929 0.907 0.930 0.927 0.929 0.925 0.902
     BrownBear 0.906 0.002 0.943 0.887 0.935 0.906 0.944 0.915 0.939 0.940 0.875 0.872 0.910 0.934 0.930 0.936 0.938 0.937 0.269 0.940 0.935 0.936 0.923 0.915
          Carp 0.943 0.943 0.006 0.946 0.954 0.947 0.955 0.952 0.951 0.957 0.949 0.950 0.952 0.956 0.946 0.956 0.953 0.954 0.945 0.960 0.950 0.953 0.942 0.960
           Cat 0.897 0.887 0.946 0.003 0.926 0.897 0.942 0.905 0.928 0.931 0.870 0.872 0.885 0.919 0.922 0.933 0.932 0.931 0.885 0.929 0.920 0.934 0.919 0.897
    Chimpanzee 0.925 0.935 0.954 0.926 0.006 0.926 0.948 0.926 0.849 0.731 0.925 0.922 0.921 0.943 0.667 0.943 0.841 0.946 0.931 0.441 0.933 0.835 0.934 0.930
           Cow 0.883 0.906 0.947 0.897 0.926 0.006 0.936 0.885 0.931 0.927 0.890 0.888 0.893 0.925 0.920 0.931 0.930 0.929 0.905 0.931 0.921 0.930 0.923 0.899
       Echidna 0.936 0.944 0.955 0.942 0.948 0.936 0.005 0.936 0.947 0.947 0.940 0.937 0.942 0.941 0.939 0.936 0.947 0.855 0.935 0.949 0.941 0.947 0.929 0.948
  FinbackWhale 0.616 0.915 0.952 0.905 0.926 0.885 0.936 0.005 0.930 0.931 0.911 0.908 0.901 0.933 0.922 0.936 0.933 0.934 0.910 0.932 0.928 0.932 0.927 0.902
        Gibbon 0.928 0.939 0.951 0.928 0.849 0.931 0.947 0.930 0.005 0.859 0.932 0.930 0.927 0.948 0.844 0.951 0.872 0.952 0.936 0.854 0.939 0.868 0.933 0.929
       Gorilla 0.931 0.940 0.957 0.931 0.731 0.927 0.947 0.931 0.859 0.006 0.927 0.929 0.924 0.944 0.737 0.944 0.835 0.943 0.928 0.732 0.938 0.836 0.934 0.929
      GreySeal 0.901 0.875 0.949 0.870 0.925 0.890 0.940 0.911 0.932 0.927 0.003 0.399 0.888 0.924 0.922 0.933 0.931 0.936 0.863 0.929 0.922 0.930 0.920 0.898
    HarborSeal 0.898 0.872 0.950 0.872 0.922 0.888 0.937 0.908 0.930 0.929 0.399 0.004 0.888 0.922 0.922 0.933 0.932 0.937 0.860 0.930 0.922 0.928 0.919 0.900
         Horse 0.896 0.910 0.952 0.885 0.921 0.893 0.942 0.901 0.927 0.924 0.888 0.888 0.003 0.928 0.913 0.937 0.923 0.936 0.903 0.923 0.912 0.924 0.924 0.848
    HouseMouse 0.926 0.934 0.956 0.919 0.943 0.925 0.941 0.933 0.948 0.944 0.924 0.922 0.928 0.006 0.932 0.923 0.944 0.930 0.924 0.942 0.860 0.945 0.921 0.928
         Human 0.920 0.930 0.946 0.922 0.667 0.920 0.939 0.922 0.844 0.737 0.922 0.922 0.913 0.932 0.005 0.949 0.834 0.949 0.931 0.681 0.938 0.826 0.934 0.929
       Opossum 0.936 0.936 0.956 0.933 0.943 0.931 0.936 0.936 0.951 0.944 0.933 0.933 0.937 0.923 0.949 0.006 0.960 0.938 0.939 0.954 0.941 0.960 0.891 0.952
     Orangutan 0.928 0.938 0.953 0.932 0.841 0.930 0.947 0.933 0.872 0.835 0.931 0.932 0.923 0.944 0.834 0.960 0.006 0.954 0.933 0.843 0.943 0.585 0.945 0.934
      Platypus 0.929 0.937 0.954 0.931 0.946 0.929 0.855 0.934 0.952 0.943 0.936 0.937 0.936 0.930 0.949 0.938 0.954 0.003 0.932 0.948 0.937 0.949 0.920 0.948
     PolarBear 0.907 0.269 0.945 0.885 0.931 0.905 0.935 0.910 0.936 0.928 0.863 0.860 0.903 0.924 0.931 0.939 0.933 0.932 0.002 0.942 0.940 0.936 0.927 0.917
    PygmyChimp 0.930 0.940 0.960 0.929 0.441 0.931 0.949 0.932 0.854 0.732 0.929 0.930 0.923 0.942 0.681 0.954 0.843 0.948 0.942 0.007 0.935 0.838 0.931 0.929
           Rat 0.927 0.935 0.950 0.920 0.933 0.921 0.941 0.928 0.939 0.938 0.922 0.922 0.912 0.860 0.938 0.941 0.943 0.937 0.940 0.935 0.006 0.939 0.922 0.922
  SumOrangutan 0.929 0.936 0.953 0.934 0.835 0.930 0.947 0.932 0.868 0.836 0.930 0.928 0.924 0.945 0.826 0.960 0.585 0.949 0.936 0.838 0.939 0.007 0.942 0.937
      Wallaroo 0.925 0.923 0.942 0.919 0.934 0.923 0.929 0.927 0.933 0.934 0.920 0.919 0.924 0.921 0.934 0.891 0.945 0.920 0.927 0.931 0.922 0.942 0.005 0.935
    WhiteRhino 0.902 0.915 0.960 0.897 0.930 0.899 0.948 0.902 0.929 0.929 0.898 0.900 0.848 0.928 0.929 0.952 0.934 0.948 0.917 0.929 0.922 0.937 0.935 0.002
\end{verbatim}
}
\caption{Distance matrix of pairwise NCD. For display purpose,
we  have truncated the original 
entries from 15 decimals to 3 decimals precision.}\label{fig.distmatr}
\end{figure}
\end{center}
\subparagraph{Hierarchical versus Flat Clustering:}
This is a good place to contrast the informativeness of 
hierarchical clustering with multidimensional clustering 
using the same NCD matrix, exhibited in
Figure~\ref{fig.distmatr}. The entries give a good example of typical
NCD values; we truncated the number of decimals
from 15 to 3 significant digits to save space.
Note that the majority of distances bunches in the range $[0.9,1]$. This is due
to the regularities the compressor can perceive.
The diagonal elements give the self-distance, which, for PPMZ, is not
actually 0, but is off from 0 only in the third decimal.
In Figure~\ref{fig.mammal2d} we clustered the 24 animals using the NCD matrix
by multidimenional scaling as points in 2-dimensional Euclidean space.
In this method,
the NCD matrix of 24 animals can be viewed as a set of 
distances between points in $n$-dimensional Euclidean 
space ($n \leq 24$), which
we want to project
into a 2-dimensional Euclidean space, trying to distort the distances
between the pairs
as little as possible. This is akin to the problem of projecting
the surface of the earth globe on a two-dimensional map with minimal
distance distortion. The main feature is the choice of the measure
of distortion to be minimized,  \cite{DHS}. Let the original set of distances
be $d_1, \ldots , d_k$ and the projected distances be
$d'_1, \ldots , d'_k$. 
 In Figure~\ref{fig.mammal2d} we used
the distortion measure {\em Kruskall's stress-1}, 
\cite{Kr64}, which minimizes 
$\sqrt{(\sum_{i \leq k} (d_i -d'_i)^2)/ \sum_{i \leq k} d_i^2}$.
Kruskall's stress-1 equal 0 means no distortion, and the worst value
is at most 1 (unless you have a really bad projection).
In the projection of the NCD matrix according to our quartet method
one minimizes the more subtle distortion $S(T)$ measure, where 1 means
perfect representation of the relative relations between every 4-tuple,
and 0 means minimal representation.  
Therefore, we should compare distortion Kruskall stress-1 with $1-S(T)$.
Figure~\ref{tree-mammal} has a very good
$1-S(T)=0.04$ and
Figure~\ref{fig.mammal2d} has a poor Kruskal stress $0.389$. 
Assuming that the comparison is significant for small values (close to perfect
projection), we find that 
the multidimensional scaling of this experiment's NCD matrix
is formally inferior to that of the quartet
tree. This conclusion formally justifies
the impression conveyed by the figures on visual inspection.

\begin{figure}
\hfill\ \psfig{figure=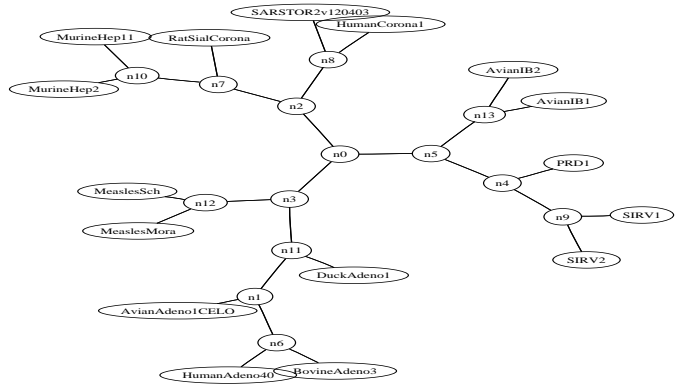,width=3.5in,height=2in} \hfill\
\caption{SARS virus among other virii. Legend:
AvianAdeno1CELO.inp:  Fowl adenovirus 1;
AvianIB1.inp:  Avian infectious bronchitis virus (strain Beaudette US);
AvianIB2.inp:  Avian infectious bronchitis virus (strain Beaudette CK);
BovineAdeno3.inp:  Bovine adenovirus 3;
DuckAdeno1.inp:  Duck adenovirus 1;
HumanAdeno40.inp:  Human adenovirus type 40;
HumanCorona1.inp:  Human coronavirus 229E;
MeaslesMora.inp:  Measles virus strain Moraten;
MeaslesSch.inp:  Measles virus strain Schwarz;
MurineHep11.inp:  Murine hepatitis virus strain ML-11;
MurineHep2.inp:  Murine hepatitis virus strain 2;
PRD1.inp:  Enterobacteria phage PRD1;
RatSialCorona.inp:  Rat sialodacryoadenitis coronavirus;
SARS.inp: SARS TOR2v120403;
SIRV1.inp:  Sulfolobus virus SIRV-1;
SIRV2.inp:  Sulfolobus virus SIRV-2.
$S(T)=0.988$.
}\label{fig.sars}
\end{figure}

\paragraph{SARS Virus:}
In another experiment we clustered the SARS virus after its
sequenced genome was made publicly available,
 in relation to potential similar virii.
The 15 virus genomes were downloaded from The Universal Virus 
Database of the International Committee on Taxonomy of Viruses,
available on the world-wide web.  
The SARS virus was downloaded from 
Canada's Michael Smith Genome Sciences Centre which
had the first public SARS Coronovirus draft whole genome 
assembly available for download (SARS TOR2 draft genome assembly 120403).
The NCD distance matrix was computed using the compressor bzip2.
The relations in Figure~\ref{fig.sars} are very similar to the
definitive tree based on medical-macrobio-genomics analysis,
appearing later in the New England Journal of Medicine,
\cite{SA03}.
We depicted the figure in the ternary tree style, rather than the 
genomics-dendrogram style, since the former is more precise
for visual inspection of proximity relations.
\begin{figure}
\hfill\ \psfig{figure=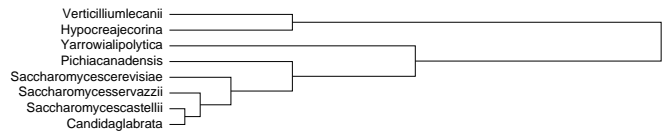, width=3.5in} \hfill\
\caption{Dendrogram of mitochondrial genomes of fungi using NCD.
This represents the distance matrix precisely with
$S(T)=0.999$.}\label{fig.fungi}
\end{figure}
\begin{figure}
\hfill\ \psfig{figure=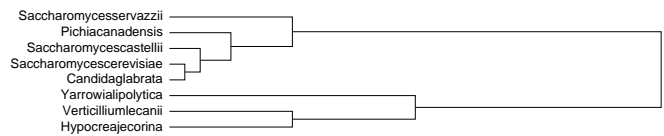, width=3.5in} \hfill\
\caption{Dendrogram of mitochondrial genomes of fungi using block frequencies.
This represents the distance matrix precisely with 
$S(T)=0.999$.}\label{fig.vbf}
\end{figure}

\paragraph{Analysis of Mitochondrial Genomes of Fungi:}
As a pilot for applications of the CompLearn Toolkit
in fungi genomics reasearch, the group of T. Boekhout,
E. Kuramae,
 V. Robert,
of the
Fungal Biodiversity Center, Royal Netherlands Academy of Sciences,
supplied us with the
mitochondrial genomes of 
{\em Candida
glabrata, Pichia canadensis, Saccharomyces cerevisiae, S. castellii, S.
servazzii, Yarrowia lipolytica} (all yeasts), and two filamentous ascomycetes
{\em Hypocrea jecorina} and {\em Verticillium lecanii}.
The NCD distance matrix was computed using the compressor PPMZ.
The resulting tree is depicted in Figure~\ref{fig.fungi}.
The interpretation of the fungi researchers is
``the tree clearly clustered the
ascomycetous yeasts versus the two filamentous Ascomycetes, 
thus supporting the current hypothesis on their classification 
(for example, see  \cite{KS01}).
Interestingly, in a recent treatment of the Saccharomycetaceae, S. servazii, 
S. castellii and C. glabrata were all proposed to belong to 
genera different from
Saccharomyces, and this is supported by the topology of our tree as well
(\cite{Ku03}).''

To compare the veracity of the NCD clustering with a more feature-based
clustering, we also calculated the pairwise distances as follows:
Each file is converted to a $4096$-dimensional vector by considering
the frequency of all (overlapping) 6-byte contiguous blocks.
The l2-distance (Euclidean distance) is calculated between each pair of
files by taking the square root of the sum of the squares of the
component-wise differences.
These distances are arranged into a distance matrix 
and linearly scaled to
fit the range $[0,1.0]$. Finally, we ran the clustering
routine on this distance matrix. The results are in Figure~\ref{fig.vbf}.
As seen by comparing with the NCD-based
Figure~\ref{fig.fungi}
there are apparent misplacements when using the Euclidean
distance in this way.  Thus, in this simple experiment, the NCD performed
better, that is, agreed more precisely with accepted biological knowledge.

\subsection{Language Trees}
                                                                                
\begin{figure}
\hfill\ \psfig{figure=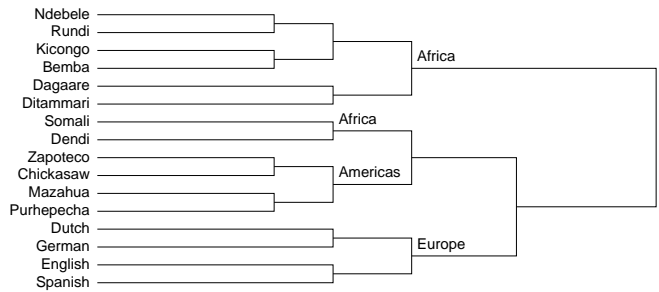, width=3.5in} \hfill\
\caption{Clustering of Native-American, Native-African,
and Native-European languages. $S(T)=0.928$.}\label{fig.lang}
\end{figure}
Our method  improves the results of
\cite{BCL02a},
 using a linguistic corpus of ``The
Universal Declaration of Human Rights (UDoHR)'' \cite{rights} in 52
languages. Previously, \cite{BCL02a} used an asymmetric measure
based on relative entropy, and the full matrix of the pair-wise
distances between all 52 languages, to
build a language
classification tree. This experiment was repeated (resulting in a somewhat
better tree) using the compression method in \cite{Li03} using standard
biological software packages to construct the phylogeny. We have 
redone this experiment, and done new experiments,
 using the CompLearn Toolkit.
Here, we report on an experiment to separate radically different
language families.
We downloaded the language versions
of the UDoHR text in English, Spanish, Dutch, German (Native-European),
Pemba, Dendi, Ndbele, Kicongo, Somali, Rundi, 
Ditammari, Dagaare (Native African),
Chikasaw, Perhupecha, Mazahua, Zapoteco (Native-American), and 
didn't preprocess them except for removing initial identifying information.
 %In each case,
%first, we transformed each UNICODE character
%in the text of the ``Universal Declaration of Human Rights''
%in the particular language being investigated
 %into an ASCII character by removing its
%vowel flag if necessary. Secondly, 
We used an Lempel-Ziv-type compressor {\it gzip}
to compress text sequences of
sizes not exceeding the length of the sliding window {\it gzip} uses
(32 kilobytes), and compute the NCD
for each pair of language sequences. Subsequently we clustered
the result. We show the outcome of one of the experiments in
Figure~\ref{fig.lang}. 
Note that three groups are correctly clustered,
and that even the subclusters of the European languages are correct
(English is grouped with the Romance languages because it contains
up to 40\% admixture of words from Latin origine).

\subsection{Literature}
\begin{figure}
\hfill\ \psfig{figure=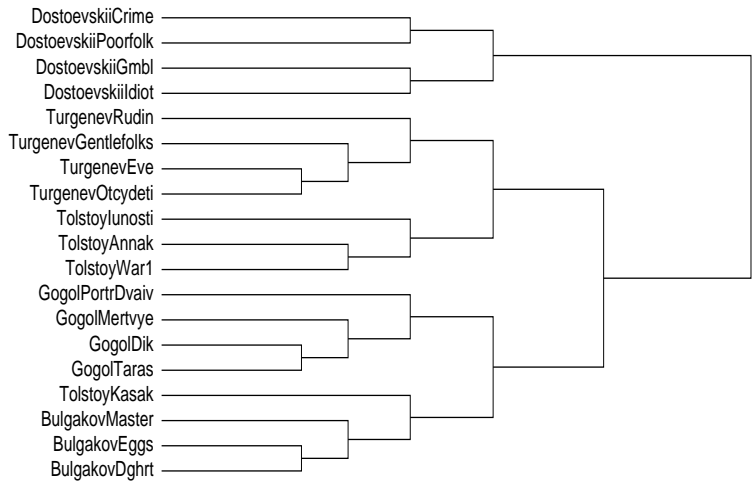,width=4in,height=2.5in} \hfill\
\caption{Clustering of Russian writers.
Legend: I.S. Turgenev, 1818--1883 [Father and Sons, Rudin, On the Eve, 
A House of Gentlefolk]; F. Dostoyevsky 1821--1881 [Crime and Punishment, 
The Gambler,
The Idiot; Poor Folk]; L.N. Tolstoy 1828--1910 [Anna Karenina, The Cossacks,
Youth, War and Piece]; N.V. Gogol 1809--1852 [Dead Souls, Taras Bulba,
The Mysterious Portrait,  How the Two Ivans Quarrelled];
M. Bulgakov 1891--1940 [The Master and Margarita, The Fatefull Eggs, The 
Heart of a Dog]. $S(T)=0.949$.
}\label{fig.russwriter}
\end{figure}

The texts used in this experiment were down-loaded from the world-wide web 
in original Cyrillic-lettered Russian and in Latin-lettered English
by L. Avanasiev (Moldavian MSc student at the University of Amsterdam).
The compressor used to compute the NCD matrix was bzip2.
We clustered Russian literature in the original 
(Cyrillic) by Gogol, Dostojevski, Tolstoy, Bulgakov,Tsjechov,
with three or four different texts per author. Our purpose was to
see whether the clustering is sensitive enough, and the authors distinctive
enough, to result in clustering by
author. In Figure~\ref{fig.russwriter} we see a perfect clustering.
\begin{figure}
\hfill\ \psfig{figure=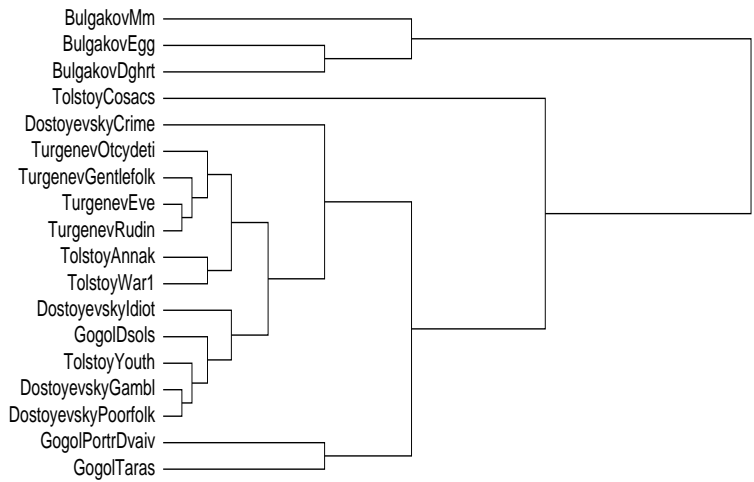,width=4in,height=2.5in} \hfill\
\caption{Clustering of Russian writers translated in English.
The translator is given in brackets after the titles of the texts.
Legend: I.S. Turgenev, 1818--1883 [Father and Sons (R. Hare), Rudin
(Garnett, C. Black), On the Eve (Garnett, C. Black),
A House of Gentlefolk (Garnett, C. Black)]; 
F. Dostoyevsky 1821--1881 [Crime and Punishment (Garnett, C. Black),
The Gambler (C.J. Hogarth),
The Idiot (E. Martin); Poor Folk  (C.J. Hogarth)]; 
L.N. Tolstoy 1828--1910 [Anna Karenina  (Garnett, C. Black), The Cossacks
(L. and M. Aylmer),
Youth (C.J. Hogarth), War and Piece (L. and M. Aylmer)]; 
N.V. Gogol 1809--1852 [Dead Souls (C.J. Hogarth), Taras Bulba ($\approx$
G. Tolstoy, 1860, B.C. Baskerville),
The Mysterious Portrait +  How the Two Ivans Quarrelled ($\approx$
I.F. Hapgood];
M. Bulgakov 1891--1940 [The Master and Margarita (R. Pevear, L. Volokhonsky), 
The Fatefull Eggs (K. Gook-Horujy), The
Heart of a Dog (M. Glenny)]. $S(T)= 0.953$.
}\label{fig.russtrans}
\end{figure}
Considering the English translations of the same texts,
in Figure~\ref{fig.russtrans}, we see errors in the clustering.
Inspection shows that the clustering is now partially based on the 
translator. It appears that the translator superimposes his characteristics
on the texts, partially suppressing the characteristics of the original
authors.
In other experiments we separated authors by gender and by period.

\subsection{Music}\label{secdetails}
The amount of digitized music available on the internet has grown 
dramatically in recent years, both in the public domain 
and on commercial sites. Napster and its clones are prime examples.
Websites offering musical content in some form or other
(MP3, MIDI, \ldots) need a way to organize their wealth of material;
they need to somehow classify their files according to 
musical genres and subgenres, putting similar pieces together.
The purpose of such organization is to enable users 
to navigate to pieces of music they already know and like, 
but also to give them advice and recommendations
(``If you like this, you might also like\ldots'').
Currently, such organization is mostly done manually by humans,
but some recent research has been looking into the possibilities 
of automating music classification.

\begin{figure}
\begin{center}
\epsfig{file=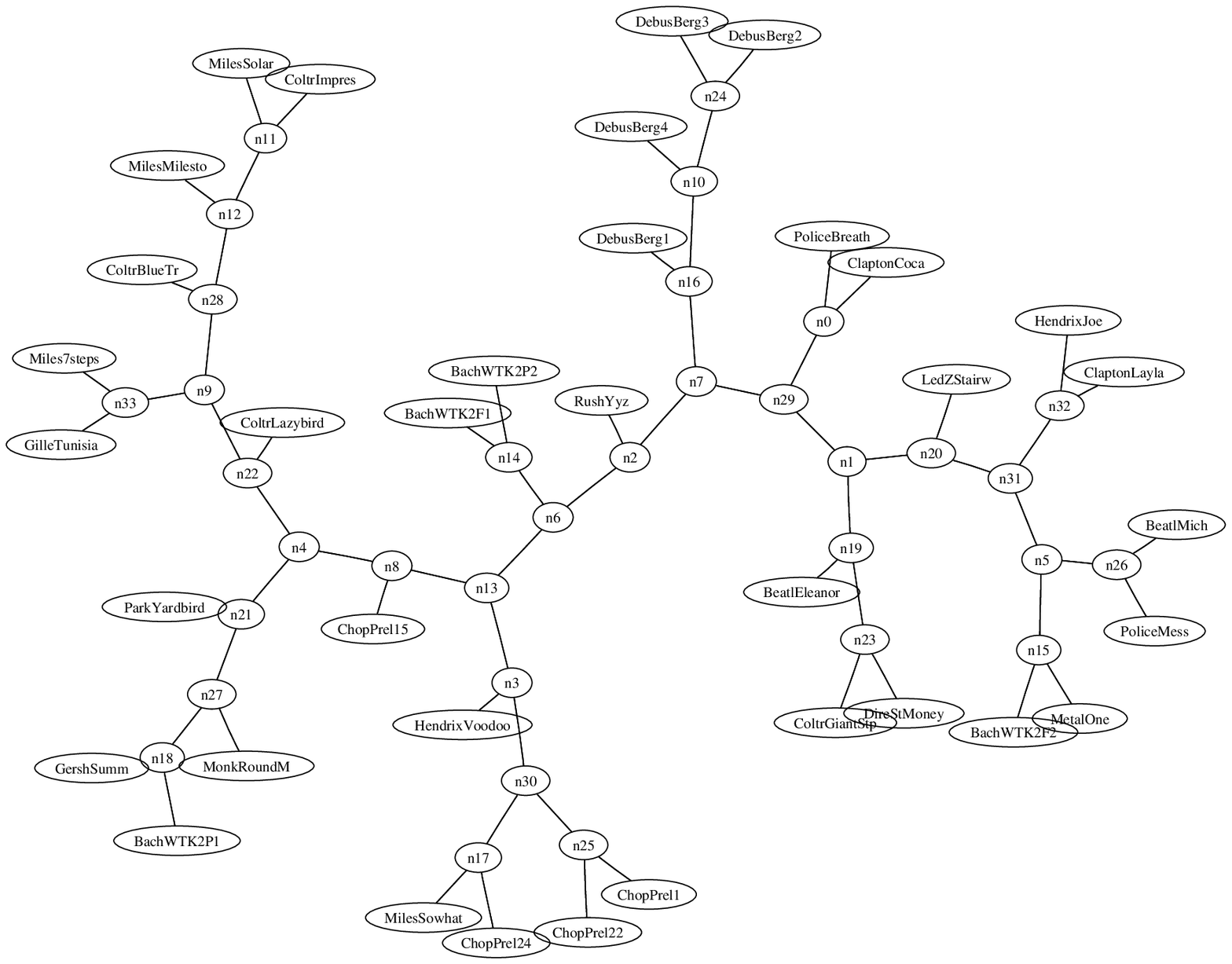,width=6in,height=4in}
\end{center}
\caption{Output for the 36 pieces from 3 music-genres. 
Legend: 12 Jazz: John Coltrane [Blue Trane,
              Giant Steps,
              Lazy Bird,
              Impressions];
Miles Davis   [Milestones,
              Seven Steps to Heaven,
              Solar,
              So What];
George Gershwin  [Summertime];
Dizzy Gillespie  [Night in Tunisia];
Thelonious Monk  [Round Midnight];
Charlie Parker   [Yardbird Suite];
12 Rock \& Pop: The Beatles   [Eleanor Rigby,
              Michelle];
Eric Clapton  [Cocaine,
              Layla];
Dire Straits  [Money for Nothing];
Led Zeppelin  [Stairway to Heaven];
Metallica     [One];
Jimi Hendrix [Hey Joe,
              Voodoo Chile];
The Police   [Every Breath You Take,
              Message in a Bottle]
Rush          [Yyz];
12 Classic: see Legend Figure~\ref{figsmallset}.
$S(T)=0.858$.}\label{figgenres}
\end{figure}
\begin{figure}
\begin{center}
\epsfig{file=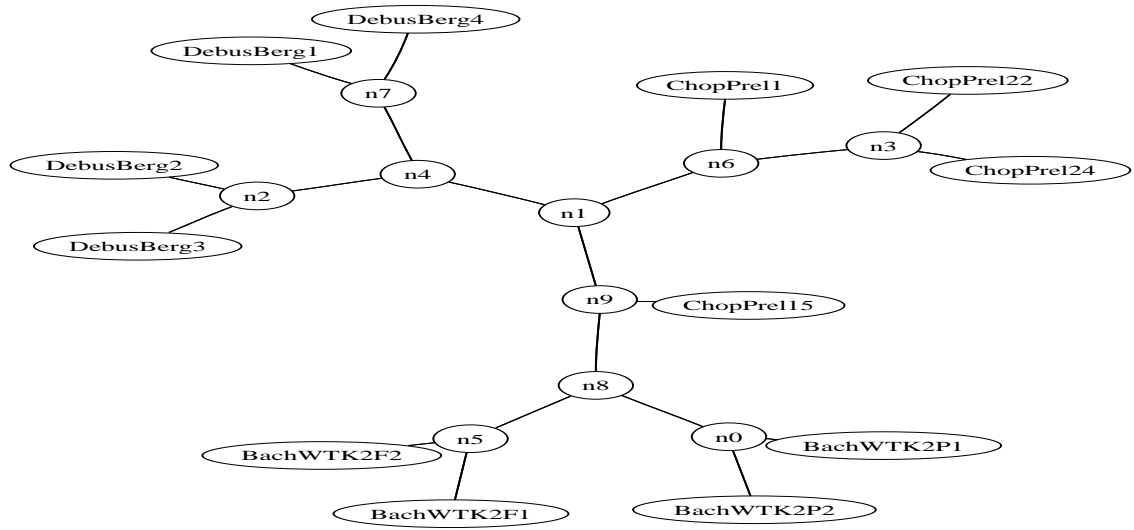,width=15cm,height=7cm}
\end{center}
\caption{Output for the 12-piece set. 
Legend: J.S. Bach [Wohltemperierte Klavier II: Preludes and Fugues 1,2---
BachWTK2\{F,P\}\{1,2\}]; Chopin [Pr\'eludes op.~28: 1, 15, 22, 24
---ChopPrel\{1,15,22,24\}]; 
Debussy [Suite Bergamasque, 4 movements---DebusBerg\{1,2,3,4\}].
$S(T)=0.968$.}\label{figsmallset}
\end{figure}
Initially, we downloaded 36 separate MIDI (Musical Instrument Digital
Interface, a versatile digital music format
available on the world-wide-web) 
files selected from a range of classical composers, as well as some
popular music.  The files were down-loaded from several different
MIDI Databases on the world-wide web. The identifying information,
composer, title, and so on, was stripped from the files (otherwise
this may give a marginal advantage to identify composers to the compressor). 
Each of these files was run through a preprocessor 
to extract just MIDI Note-On
and Note-Off events.  These events were then converted to a player-piano
style representation, with time quantized in $0.05$ second intervals.
All instrument indicators, MIDI control signals, and tempo variations were 
ignored.  For each track in the MIDI file, we calculate two quantities:
An {\em average volume} and a {\em modal note}.
The average volume is calculated by averaging the volume (MIDI note velocity)
of all notes in the track.  The modal note is defined to be the note 
pitch that sounds most often in that track.  If this is not unique, 
then the lowest such note is chosen.  The modal note is used as a 
key-invariant reference point from which to represent all notes.  
It is denoted by $0$, higher notes are denoted by positive numbers, and 
lower notes are denoted by negative numbers.  A value of $1$ indicates 
a half-step above the modal note, and a value of $-2$ indicates
a whole-step below the modal note.  The tracks are sorted according to
decreasing average volume, and then output in succession.  For each track,
we iterate through each time sample in order, outputting a single signed
8-bit value for each currently sounding note.  Two special values are
reserved to represent the end of a time step and the end of a track.  This
file is then used as input to the compression stage for distance
matrix calculation and subsequent tree search.
To check whether any important feature of the music was lost during
preprocessing, we played it back from the preprocessed files to verify
whether it sounded like the original. To the authors the pieces sounded almost
unchanged.
The compressor used to compute the NCD matrix of the genres tree,
Figure~\ref{figgenres},
and that of 12-piece music set, Figure~\ref{figsmallset} is bzip2.
For the full range of the music experiments see \cite{CVW03}.

Before testing whether our program can see the distinctions
between various classical composers, we first 
show that it can distinguish between three broader musical genres:
classical music, rock, and jazz. This may be easier than
making distinctions ``within'' classical music. 
All musical pieces we used are listed in the tables in the full paper
(on the URL provided above).
For the genre-experiment we used 12 classical pieces 
consisting of Bach, Chopin, and Debussy,
12 jazz pieces, and
12 rock pieces.
The tree (Figure~\ref{figgenres}) that our program came up with 
has $S(T)=0.858$.
\begin{figure}
\begin{center}
\epsfig{file=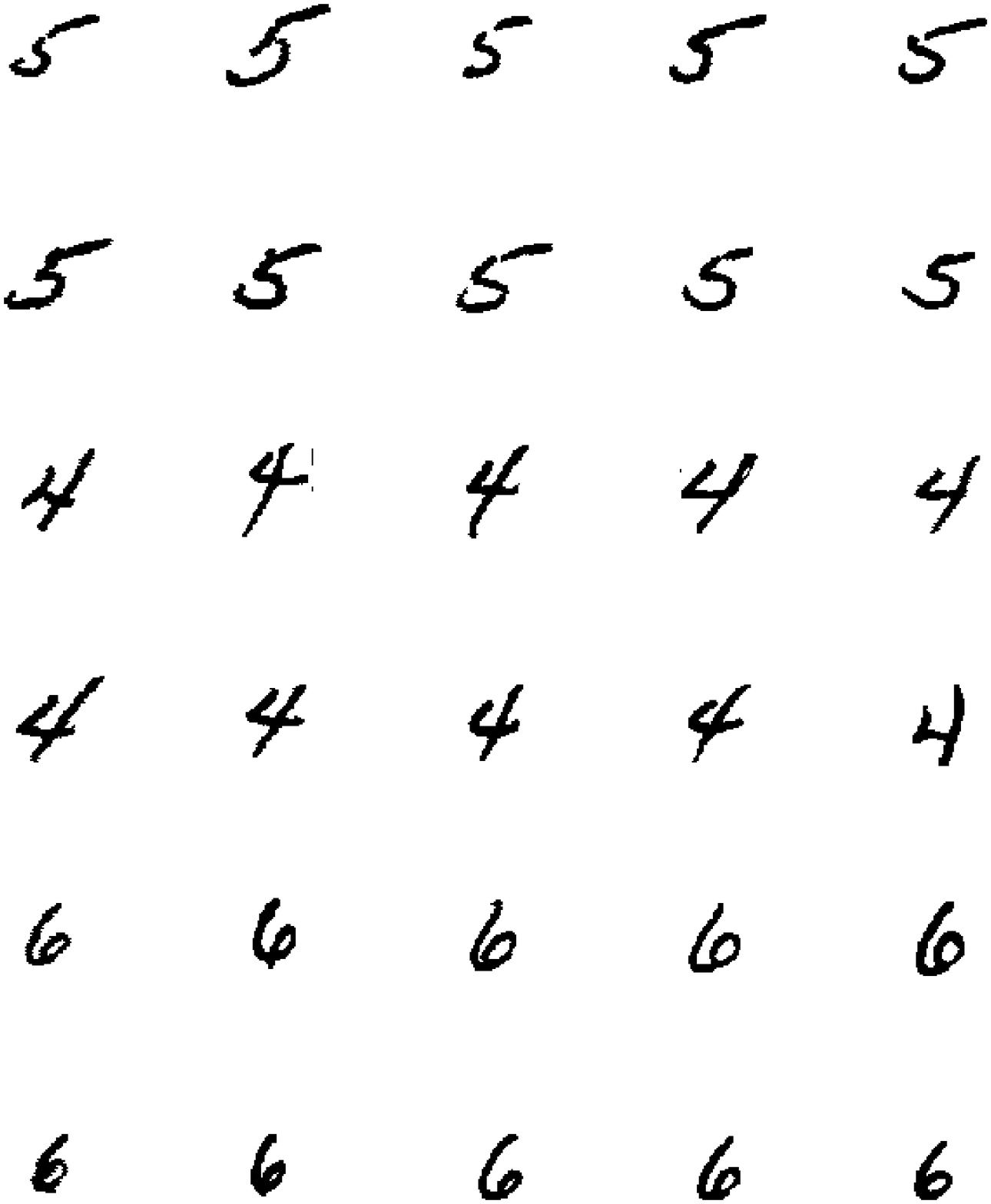,width=2.5in,height=2.5in}
\end{center}
\caption{Images of handwritten digits used for OCR.}\label{fig.ocrdata}
\end{figure}
The discrimination between the 3 genres is reasonable but not perfect.
Since $S(T)=0.858$, a fairly low value, the resulting tree doesn't
represent the NCD distance matrix very well. Presumably,
the information in the NCD distance matrix cannot be represented
by a dendrogram of high $S(T)$ score. This appears to be a common problem
with large ($>25$ or so) natural data sets. Another reason may
be that the program terminatedi, while trapped in a local optimum. We
repeated the experiment many times with almost the same results,
so that doesn't appear to be the case.
The 11-item subtree rooted at $n4$ contains 10 of the 12 jazz pieces,
together with a piece of Bach's ``Wohltemporierte Klavier (WTK)''.
The other two jazz pieces, Miles Davis' ``So What,'' and John 
Coltrane's ``Giant Steps'' %, and Gershwin's ``Summertime'',
 are placed elsewhere in the tree,
perhaps according to some kinship that now escapes us (but may be
identified by closer studying of the objects concerned).
Of the 12 rock pieces, 10 are placed in the 12-item subtree rooted at $n29$,
together with a piece of Bach's ``WTK,'' and Coltrane's ``Giant Steps,'' 
while Hendrix's ``Voodoo Chile'' and Rush ``Yyz'' is 
further away.
%In the case of the Hendrix piece this may be explained by the fact
%that it does not fit well in a specific genre.
Of the 12 classical pieces, 10 are in the 13-item subtrees rooted
at the branch $n8,n13,n6,n7$, together with Hendrix's ``Voodoo Chile,''
Rush's ``Yyz,'' and Miles Davis' ``So What.''
Surprisingly, 2 of the 4 Bach ``WTK'' pieces are placed elsewhere.
Yet we perceive the 4 Bach pieces to  be very close,
both structurally and melodically (as they all come from the mono-thematic
``Wohltemporierte Klavier''). But the program
finds a reason that at this
point is hidden from us. In fact, running this experiment with different
compressors and termination conditions consistently displayed
this anomaly. 
%However, Bach's is a seminal music and has been copied and cannibalized
%in all kinds of recognizable or hidden manners; closer scrutiny could
%reveal likenesses in its present company that are not now apparent to us.
\begin{figure}
\begin{center}
\epsfig{file=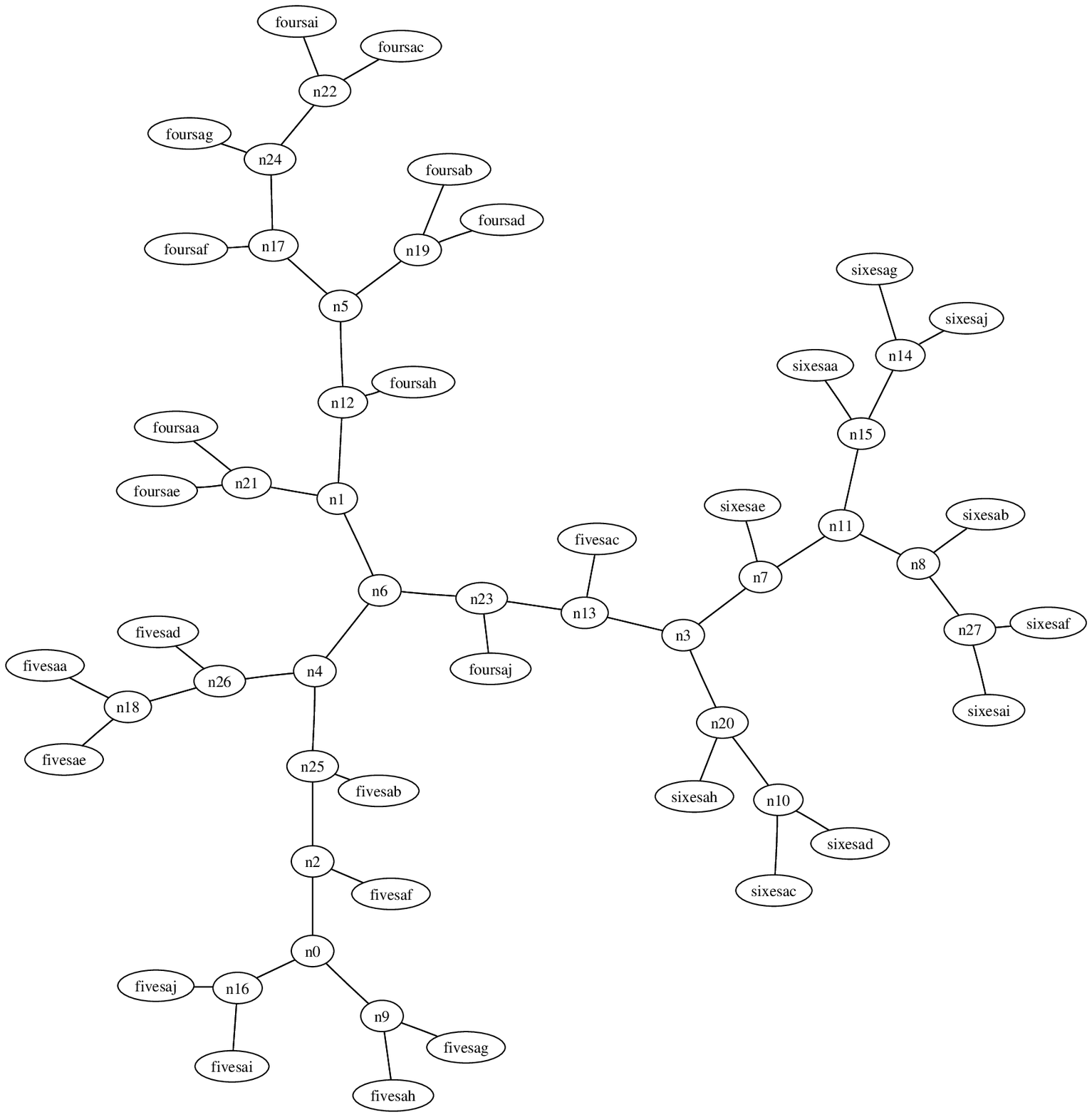,width=4in,height=3in}
\end{center}
\caption{Clustering of the OCR images. $S(T)=  0.901$.}\label{fig.ocr}
\end{figure}
The small set encompasses the 4 movements from Debussy's ``Suite Bergamasque,''
4 movements of book 2 of Bach's ``Wohltemperierte Klavier,'' 
and 4 preludes from 
Chopin's ``Opus~28.'' As one can see in Figure~\ref{figsmallset}, 
our program does a pretty good job at clustering these pieces.
The $S(T)$ score is also high: 0.968.
The 4 Debussy movements form one cluster, as do the 4 Bach pieces.  
The only imperfection in the tree, judged by what one would 
intuitively expect, is that Chopin's Pr\'elude no.~15 lies a bit closer 
to Bach than to the other 3 Chopin pieces.
This Pr\'elude no~15, in fact, consistently forms an odd-one-out 
in our other experiments as well. This is an example of pure data mining,
since there is some musical truth 
to this, as no.~15 is perceived as by far the most eccentric 
among the 24 Pr\'eludes of Chopin's opus~28.

%We further tested the method with
%a medium set that adds 20 pieces to the small set,
%which gave an $S(T)$ score is slightly lower than
%in the small set experiment: 0.895;
%a large set of 60 pieces 
%where the $S(T)$ score dropped further from that of the
%medium-sized set to 0.844;
%more complicated
%music, 34 symphonic pieces,
%which resulted in an $S(T)$ score of 0.860. In all cases the $S(T)$
%score is reliable with respect to what our intuition tells us.

\subsection{Optical Character Recognition}
\label{sec.ocr}
Can we also cluster two-dimensional images? Because our method
appears focussed on strings this is not straightforward. It turns out
that scanning a picture in raster row-major order retains enough
regularity in both dimensions for the compressor to grasp.
 A simple task along these lines is to
cluster handwritten characters. The handwritten characters
in Figure~\ref{fig.ocrdata} were downloaded from the NIST Special Data Base 19
(optical character recognition database) on the world-wide web.
 Each file in the data directory contains 1 digit image,
either a
four, five, or six.  Each pixel is a single character; 
'\#' for a black pixel, '.' for white.
Newlines are added at the end of each line.  Each character is 128x128 pixels.
The NCD matrix was computed using the compressor PPMZ.
The Figure~\ref{fig.ocr}  shows each character that is used.
There are 10 of each digit ``4,'' ``5,'' ``6,''
making a total of 30 items in this experiment.
All but one of the 4's are put in the subtree rooted at $n1$, all but
one  of the 5's are put in the subtree rooted at $n4$, and all 6's
are put in the subtree rooted at $n3$. The remaining 4 and 5
are in the branch $n23,n13$ joining $n6$ and $n3$. So 28 items
out of 30 are clustered correctly, that is, 93\%.
In this experiment we used only 3 digits. Using the full set of
decimal digits results in a lower clustering accuracy. However,
we can use the NCD
as a oblivious feature-extraction technique to convert
generic objects into finite-dimensional vectors.  We have used this
technique to train a support vector machine 
(SVM) based OCR system to classify handwritten digits
by extracting 80 distinct, ordered NCD features from each input image.
In this initial stage of ongoing research, by our oblivious
method of compression-based clustering to supply a kernel
for an SVM classifier,
we achieved a handwritten single decimal digit recognition accuracy of 85\%.
The current state-of-the-art for this problem,
after half a century of interactive feature-driven
classification research, in the upper ninety \% level \cite{OSBS02,DJT96}.
All experiments are bench marked on the standard  NIST Special Data Base 19
(optical character recognition database).

\subsection{Astronomy}
\begin{figure}
\begin{center}
\epsfig{file=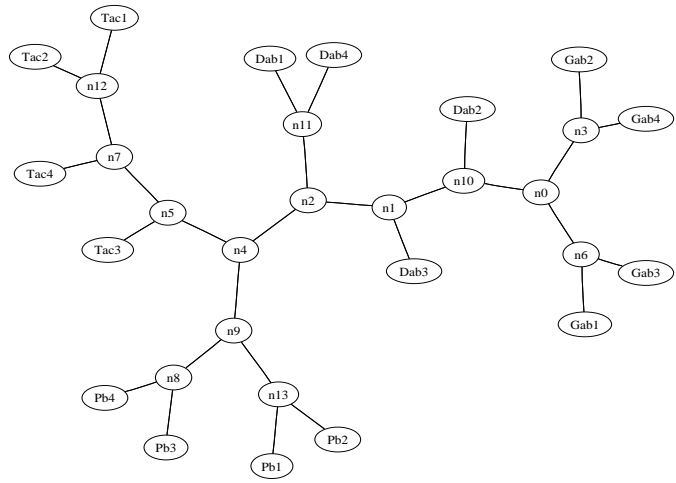,width=3.5in,height=2.5in}
\end{center}
\caption{16 observation intervals of GRS 1915+105 from four classes. 
The initial capital letter indicates the class corresponding
to Greek lower case letters in  \cite{BKWMKP00}. The remaining letters and 
digits identify the particular observation interval in terms of finer 
features and identity. The $T$-cluster is top left, the $P$-cluster
is bottom left, the $G$-cluster is to the right, and the $D$-cluster
in the middle. This tree almost exactly represents the underlying
NCD distance matrix: $S(T)= 0.994$. 
}\label{fig.astro}
\end{figure}

As a proof of principle we clustered data from unknown objects,
for example objects that are far away. 
In \cite{BKWMKP00} observations of the microquasar
GRS 1915+105 made with the Rossi X-ray Timing Eplorer were analyzed.
The interest in this microquasar stems from the
fact that it was the first Galactic object to show a certain behavior
(superluminal expansion in radio observations).
Photonometric observation data from X-ray telescopes were divided into
short time segments (usually in the order of one minute), and these segments
have been classified into a bewildering array of fifteen different
modes after considerable effort.  Briefly, spectrum
hardness ratios (roughly, ``color'') and photon count sequences were used to
classify a given interval into categories of  variability modes.
From this analysis, the extremely complex variability 
of this source was reduced to transitions between three basic states,
which, interpreted in astronomical terms, gives rise to an explanation
of this peculiar source in standard black-hole theory.
The data we used in this experiment
made available to us by M. Klein Wolt (co-author of the above paper)
and T. Maccarone, both researchers at the
Astronomical Institute ``Anton Pannekoek'', University of Amsterdam.
The observations are essentially
time series, and our aim was experimenting with our method
as a pilot to more extensive joint research. Here the task
was to see whether the clustering
would agree with the classification above. 
The NCD matrix was computed using the compressor PPMZ.
The results are in
Figure~\ref{fig.astro}. 
We clustered 12 objects, consisting of three intervals 
from four different categories denoted as
 $\delta, \gamma, \phi, \theta$ in Table 1 
of \cite{BKWMKP00}. In Figure~\ref{fig.astro}
we denote the categories by the corresponding Roman letters 
D, G, P, and T, respectively. 
The resulting tree groups these different modes 
together in a way that is consistent
with the classification by experts for these observations. 
The oblivious compression clustering corresponds
precisely with the
laborious feature-driven classification in \cite{BKWMKP00}.
                                                                                
\section{Conclusion}
To interpret what the NCD is doing, and to explain its remarkable
accuracy and robustness across application fields and compressors, 
the intuition is that
the NCD minorizes all similarity metrics based on features that are
captured by the reference compressor involved. Such features must
be relatively {\em simple} in the sense that they are expressed
by an aspect that the compressor analyzes (for example frequencies,
matches, repeats). Certain
sophisticated features may well be expressible as combinations
of such simple features, and are therefore themselves 
simple features in this sense. The extensive experimenting above shows
that even elusive features 
are captured. 

A potential application of our non-feature (or rather, many-unknown-feature)
approach is exploratory. Presented with data for which the features 
are as yet unknown, certain dominant features governing similarity
are automatically discovered by the NCD. Examining the data underlying
the clusters may yield this hitherto unknown dominant feature. 

Our experiments indicate that the NCD has application in two new areas of
support vector machine (SVM) based learning.  
Firstly, we find that the inverted NCD (1-NCD) is useful
as a kernel for generic objects in SVM learning.  Secondly,
we can use the normal NCD as a feature-extraction technique to convert
generic objects into finite-dimensional vectors, see the last
paragraph of Section~\ref{sec.ocr}.
In effect our similarity engine aims at the ideal of a perfect
data mining process, discovering unknown features in which the
data can be similar. This is the subject of current joint research 
in genomics of fungi, clinical molecular genetics, and 
radio-astronomy.  

\section*{Acknowledgement}
We thank Loredana Afanasiev, Graduate School of Logic, 
University of Amsterdam;
 Teun Boekhout, 
Eiko Kuramae,
 Vincent Robert, 
Fungal Biodiversity Center, Royal Netherlands Academy of Sciences;
Marc Klein Wolt, 
Thomas Maccarone, 
Astronomical Institute ``Anton Pannekoek'', University of Amsterdam;
Evgeny Verbitskiy, Philips Research; 
Steven de Rooij, 
Ronald de Wolf, CWI;
the referees and the editors,
for suggestions, comments, help with experiments, and data;
Jorma Rissanen and Boris Ryabko for discussions, Tzu-Kuo Huang
for pointing out some typos and simplifications, and 
Teemu Roos and Henri Tirry for implementing a visualization
of the clustering process.

\begin{small}

\end{small}

\end{document}